\documentclass[letterpaper, 10 pt, conference, usenames, dvipsnames]{ieeeconf}  % Comment this line out if you need a4paper

\IEEEoverridecommandlockouts                              % This command is only needed if 
\usepackage[dvipdfmx]{graphicx}
\graphicspath{{figs/}}
\usepackage{amsmath}
\usepackage{amssymb}
\usepackage{algorithm}
\usepackage{algpseudocode}
\usepackage{times}
\usepackage[position=bottom]{subfig}
\usepackage{xcolor}
\usepackage{hyperref}
\hypersetup{colorlinks=true}
\usepackage[space]{cite}
\usepackage{multirow}
\newcommand{\figlab}[1]{\label{figure:#1}}

\newcommand{\figref}[1]{Figure \ref{figure:#1}}
\newcommand{\tabref}[1]{Table \ref{table:#1}}

\title{\LARGE \bf
Joint Learning of Instance and Semantic Segmentation\\
for Robotic Pick-and-Place with Heavy Occlusions in Clutter
}

\author{Kentaro Wada, Kei Okada and Masayuki Inaba% <-this % stops a space
  \\
  University of Tokyo, JSK Laboratory
  \\
  \{wada, k-okada, inaba\}@jsk.imi.i.u-tokyo.ac.jp
}

\begin{document}

\maketitle
\thispagestyle{empty}
\pagestyle{empty}

\begin{abstract}
We present joint learning of instance and semantic segmentation
for visible and occluded region masks.
Sharing the feature extractor with instance occlusion segmentation,
we introduce semantic occlusion segmentation into the instance segmentation model.
This joint learning fuses the instance- and image-level reasoning
of the mask prediction on the different segmentation tasks,
which was missing in the previous work of learning instance segmentation only (instance-only).
In the experiments, we evaluated the proposed joint learning 
comparing the instance-only learning on the test dataset.
We also applied the joint learning model to 2 different types of robotic pick-and-place tasks
(random and target picking)
and evaluated its effectiveness to achieve real-world robotic tasks.

% While training, this joint learning fuses
% the instance and image-level reasoning on the mask prediction in the 2 tasks.

% Extending our previous work of learning instance occlusion segmentation,
% we introduce semantic occlusion segmentation to the instance segmentation model,
% sharing the feature extractor in both segmentation tasks.
% In this joint learning, the model is trained with
% both instance- and semantic-level reasoning for pixel-wise mask prediction.

\end{abstract}

\section{Introduction}

% ## Global problem
Recently, with the help of deep convolutional networks, the vision community has been rapidly improved
the performance of pixel-wise object segmentation with image:
{\it semantic segmentation} (predicting class label for pixels) \cite{Long:etal:CVPR2015,Chen:etal:PAMI2018,Zhao:etal:CVPR2017} and
{\it instance segmentation} (predicting class label and pixel-wise mask for instances) \cite{Dai:etal:ECCV2016,Dai:etal:NIPS2016,Li:etal:CVPR2017,He:etal:ICCV2017}.
However, these tasks have been tackled independently,
and the effect of the joint learning and collaboration of both tasks is less explored.

For robotic manipulation, pixel-wise object segmentation is a crucial component.
Previous work utilizes semantic segmentation models for pick-and-place of various objects
\cite{Jonschkowski:etal:IROS2016,Wada:etal:IROS2017,Zeng:etal:ICRA2017,Schwarz:etal:ICRA2017,Zeng:etal:ICRA2018,Schwarz:etal:ICRA2018}.
Since semantic segmentation can not segment different instances in the same class,
the work assumes that same class objects are not closely located and can be segmented
by clustering.

Our previous work \cite{Wada:etal:IROS2018} applies instance segmentation model
to instance-level picking task,
in order to overcome that limitation.
In addition to instance-level segmentation, we tackled occlusion segmentation of each instance
({\it instance occlusion segmetation})
in order to understand the stacking order of objects and decide the object picking order.
In our previous work, we extended the state of the art model of instance segmentation \cite{He:etal:ICCV2017},
that firstly detects bounding boxes of object instance and then
predicts pixel-wise mask of both visible and occluded regions inside each bounding box.
Compared to instance segmentation, instance occlusion segmentation is still challenging because the model needs
to predict bounding boxes of whole (visible + occluded) region and
occluded region mask based on the whole shape estimation in each bounding box.
Since mask prediction depends on the predicted boxes,
the predicted mask in small boxes can be truncated.

\begin{figure}[t]
  \setlength{\belowcaptionskip}{-20pt}
  \centering
  \includegraphics[width=\linewidth]{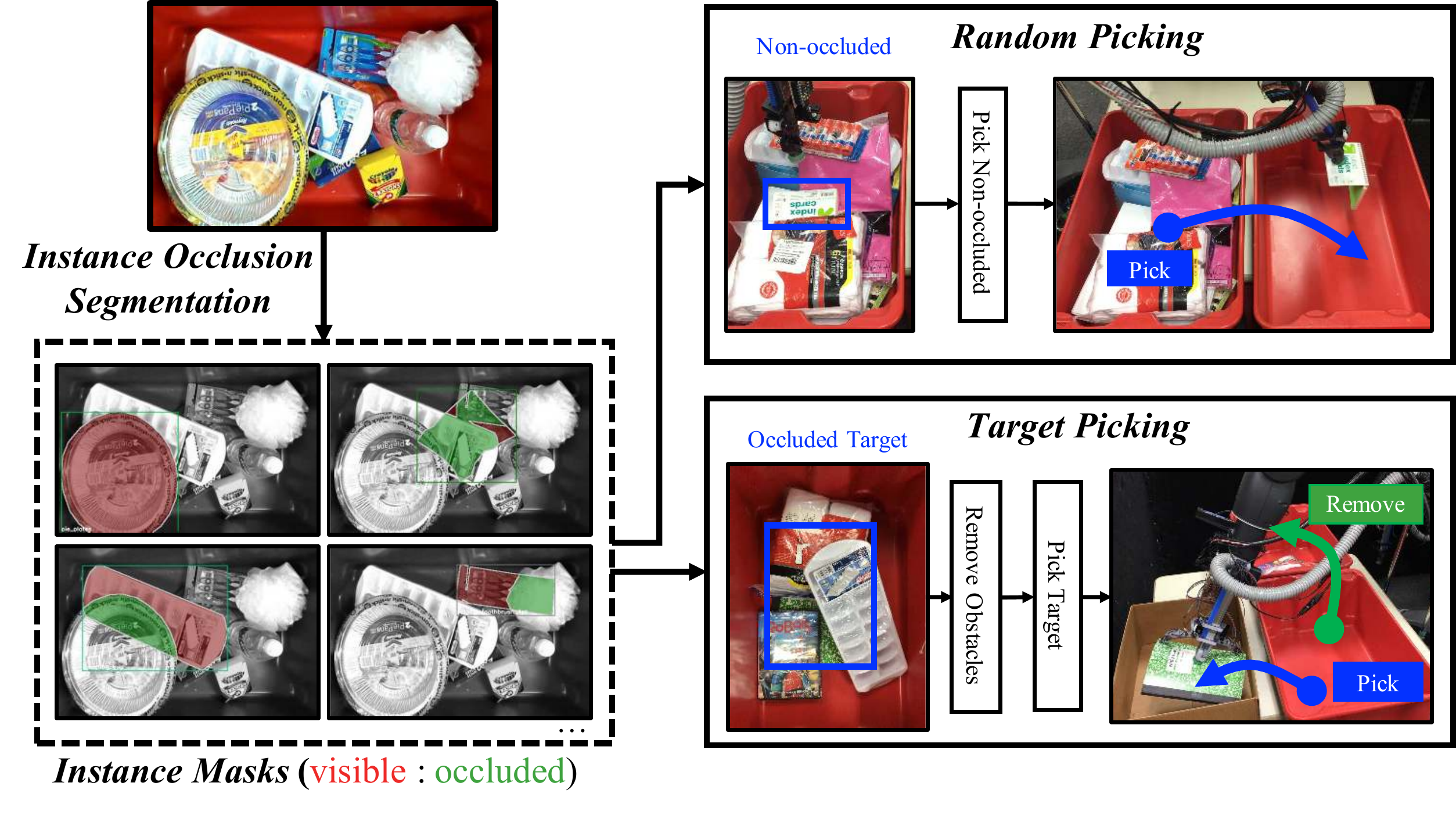}
  \caption{
    Robotic pick-and-place based on instance occlusion segmentation.
    {\footnotesize
      The instance occlusion segmentation model segments
      both visible and occluded regions of each object instance.
      This segmentation is helpful for different types of pick-and-place task:
      finding fully visible objects for {\it random picking}; and
      finding obstacle objects, which is occluding the target, for {\it target picking}.
    }
  }
  \label{figure:overview}
\end{figure}

Although instance occlusion segmentation is challenging,
it is useful for robotic pick-and-place applications: random and target picking (\figref{overview}).
For random picking in which there is no designated object,
robot needs to find non-occluded (fully visible) objects
to avoid grasp fail because of the collision to other objects.
For target picking,
robot needs to find heavily occluded target object and understand occlusion relationship among objects
for planning of the appropriate grasp order to quickly remove obstacles and access the target.
This consideration motivates us to improve instance occlusion segmentation models,
to achieve robotic picking task with heavily occluded objects in clutter.

\begin{figure*}[t]
  \setlength{\belowcaptionskip}{-10pt}
  \centering
  \includegraphics[width=\textwidth]{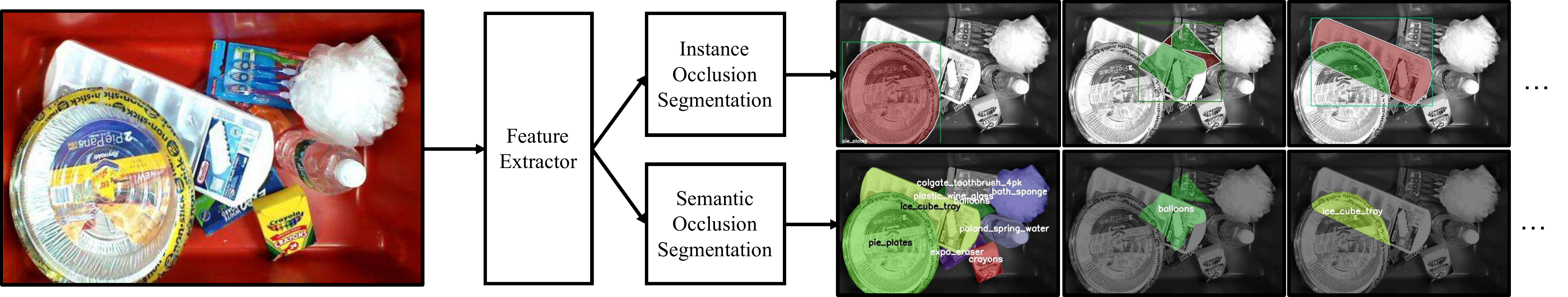}
  \caption{Joint learning of instance and semantic occlusion segmentation.}
  \label{figure:joint_learning}
\end{figure*}

In this paper, we explore the collaboration of semantic and instance occlusion segmentation
as shown in \figref{joint_learning}.
As noted before, the difficulty in instance occlusion segmentation is
caused by the two-stage prediction: $image \rightarrow box \rightarrow mask$,
especically wrong prediction of the whole bounding box in the 1st stage ($image \rightarrow box$).
On the other hand, semantic occlusion segmentation is
one-stage prediction of $mask$ from $image$: $image \rightarrow mask$,
though multiple instances in the same class are not discriminated.
We anticipate that predicting whole bounding box in instance segmentation
is difficult because there is no supervision of visible and occluded region at the 1st stage.
This motivates us to jointly train bounding box prediction and mask prediction in the 1st stage
by introducing semantic occlusion segmentation into the instance occlusion segmentation model.

% In the conventional instance segmentation model,
% the 1st stage is pixel-wise bounding box regression which is kind of {\it image-level reasoning},
% and 2nd stage is regression in the box (instance) which is {\it instance-level reasoning}.
% Since the 1st stage of semantic segmentation model is mask prediction based on image reasoning,
% the integration of these 1st stages introduces image-level reasoning of mask prediction
% which was missing in the conventional model, and we hypothesize which has a good effect for training.

% For occluded region segmentation by semantic segmentation, we extend the conventional model
% of semantic segmentation \cite{Long:etal:CVPR2015}.
% With deep convolutional networks, semantic segmentation has been achieved by
% the pixel-wise score regression and the assignment of the top-scored class to each pixel.
% This model is trained with softmax cross entropy to make it competitive among class scores,
% and the operation extracts the visible region masks of each object class.
% However, occluded region mask can not be represented in the same way 
% because there can be overlap among occluded region masks.
% To allow the overlap among occluded region masks, we train the pixel-wise score regression
% with sigmoid cross entropy, and get the occluded region mask by thresholding of the score.

In the experiments, we evaluated the proposed models
trained with a dataset which contains scenes of various objects located in clutter.
We also applied the joint learning model to real-world robotic picking task,
and demonstrated its efficiency for both target picking (pick-and-place of designated objects)
and random picking (pick-and-place of all objects) task in clutter.

% Contributions
In summary, our main contributions are:
\begin{itemize}
  \item Occluded region segmentation learning with CNN-based pixel-wise score regression;
  \item Joint learning of instance and semantic segmentation for instance visible and occluded regions;
  \item Robotic target and random picking task achievement with heavy occlusions among objects.
  % \item Achievement of robotic picking of heavily occluded target object.
\end{itemize}

\section{Related Work}

\subsection{Instance Visible and Occluded Region Segmentation}

Instance segmentation is aimed at predicting object class and
the pixel-wise mask of each instance in an image.
In previous work, this task is mainly tackled in two different approaches:
generate instance mask proposals and then classify \cite{Pinheiro:etal:NIPS2015,Pinheiro:etal:ECCV2016},
detect instances with bounding box and then apply pixel-wise segmentation \cite{Dai:etal:ECCV2016,He:etal:ICCV2017}.
Recently, Mask R-CNN \cite{He:etal:ICCV2017},
which uses the second approach is proposed as the state-of-the-art model of instance segmentation.
This model extends the bounding box detection model, Faster R-CNN~\cite{Ren:etal:NIPS2015},
to detect box and predict pixel-wise mask inside the box.
Our previous work \cite{Wada:etal:IROS2018} extends Mask R-CNN
for both visible and occluded region mask segmentation
with pixel-wise prediction of multi-class instance masks (background, visible and occluded).
% It also proposes a module for learning inter-instance relationship.
% That inter-instance relationship module is considered
% as important for occluded region segmentation
% because the occluded region of
% an instance is caused by the visible region of others.

In this paper, we introduce joint learning of instance and semantic segmentation
for visible and occluded regions.
In the semantic segmentation part, we use a similar architecture as FCIS \cite{Dai:etal:ECCV2016},
which predicts position-sensitive masks (e.g., right-top of an instance)
as a pixel-wise classification.
In the instance segmentation part, we use the extended Mask R-CNN for multi-class instance masks.
This joint learning introduces collaboration of different level of mask reasoning
of instance segmentation (instance-level reasoning for instance mask prediction)
and semantic segmentation (image-level reasoning for image pixel-wise class prediction),
which is missing in the previous model.

% of each prediction: instance-level for instance segmentation and
% image-level for semantic segmentation.

% Some work \cite{Pinheiro2015,Dai:etal:ECCV2016} tackles instance segmentation
% with firstly proposing instance masks and classification afterwards,
% which is usually called sequential approach.
% Recently, concurrent approach which simultaneously
% segments and classifies instance mask is proposed.

% Instance occlusion segmentation is an extension of instance segmentation which has been tackled
% as the task of predicting object class and pixel-wise mask of each instance in image.
% Some work \cite{Pinheiro2015,Dai:etal:ECCV2016} tackles this problem
% with firstly proposing instance masks and classification afterwards.

% Instance segmentation is aimed at predicting object class and
% the visible mask of each instance in a image.
% Instance occlusion segmentation is an extension of instance segmentation
% proposed in \cite{Wada:etal:IROS2018}.
% It is a compound task of bounding box instance detection
% and pixel-wise segmentation.
% Conventionally, it has been tackled with sequential approach
% that proposes region mask candidates and classifies them
% afterwards~\cite{Pinheiro2015,Dai:etal:ECCV2016}.
% Recently, some work \cite{Li:etal:CVPR2017,He:etal:ICCV2017} proposes

\subsection{Joint Learning}

Joint learning of different vision tasks
has been tackled in a lot of previous work.
For example, Mousavian et al. \cite{Mousavian:etal:3DV2016}
propose jointly training semantic segmentation and depth estimation
with pixel-wise score regression for image.
Cheng et al. \cite{Cheng:etal:ICCV2017} propose joint learning of
semantic segmentation and optical flow for video,
and Baslamisli et al. \cite{Baslamisli:etal:2018} propose that of
semantic segmentation and intrinsic image.
The previous work trains pixel-wise score regression model (image-level reasoning)
for different kinds of output labels (e.g., semantic labels + depth).

In the joint learning of instance and semantic segmentation,
the model is trained for very similar outputs: object region masks.
On the other hand, these outputs are predicted based on different kind of reasoning:
image-level for semantic segmentation and instance-level for instance segmentation,
which was missing in the previous work.
Also, the joint learning in this paper
does not require any additional labels annotated by human,
since ground truth semantic segmentation masks can be generated
by the masks of instance segmentation.

% As the concurrent jwork,
% joint learning of instance and semantic segmentation
% is recently proposed \cite{DeGeus2018}.
% It jointly trains instance and semantic segmentation
% for their joint task, panoptic segmentation \cite{Kirillov:etal:ARXIV2018}.
% Basically, our proposed joint learning model is similar to \cite{DeGeus2018},

\section{Joint Learning of Instance and Semantic Occlusion Segmentation}

For collaboration of instance-level reasoning of instance segmentation and
image-level reasoning of semantic segmentation,
we train a neural network model for both tasks.
As shown in \figref{joint_learning}, the model shares feature extractor
to learn commonly effective feature extraction for instance and semantic segmentation.
In the following, we describe the detail of instance and semantic occlusion segmentation model
for joint training.

\subsection{Instance Occlusion Segmentation}

% \subsubsection*{Extending Instance Visible Segmentation Model}

As in our previous work \cite{Wada:etal:IROS2018},
we extend an existing instance segmentation model, Mask R-CNN \cite{He:etal:ICCV2017},
for occlusion segmentation.
The original Mask R-CNN is designed for segmenting only the visible regions of instances,
so we extended the part of mask prediction for multi-class: visible, occluded and background
as shown in \figref{multi_class_masks}.
The model firstly predicts bounding boxes of each instance and
secondly predicts pixel-wise masks inside each box.

\begin{figure}[t]
  \setlength{\belowcaptionskip}{-10pt}
  \centering
  \begin{tabular}{c}
    \subfloat[A scene.]{
      \includegraphics[width=0.80\linewidth]{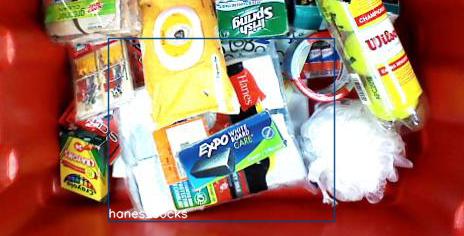}
      \figlab{multi_class_masks_image}
    }
    \\
    \subfloat[Visible.]{
      \includegraphics[width=0.25\linewidth]{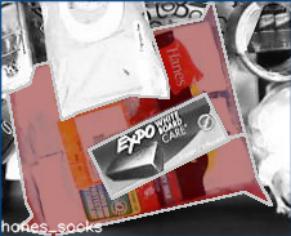}
      \figlab{multi_class_masks_visible}
    }
    \subfloat[Occluded.]{
      \includegraphics[width=0.25\linewidth]{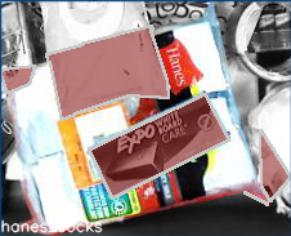}
      \figlab{multi_class_masks_occluded}
    }
    \subfloat[Background.]{
      \includegraphics[width=0.25\linewidth]{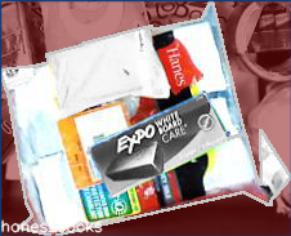}
      \figlab{multi_class_masks_background}
    }
  \end{tabular}
  \caption{Multi-class masks of an object instance.
    % {\footnotesize
    %   Each object instance has 3 different classes of regions (visible, occluded and background).
    % }
  }
  \figlab{multi_class_masks}
\end{figure}

\begin{figure}[htbp]
  \setlength{\belowcaptionskip}{-10pt}
  \centering
  \includegraphics[width=0.9\linewidth]{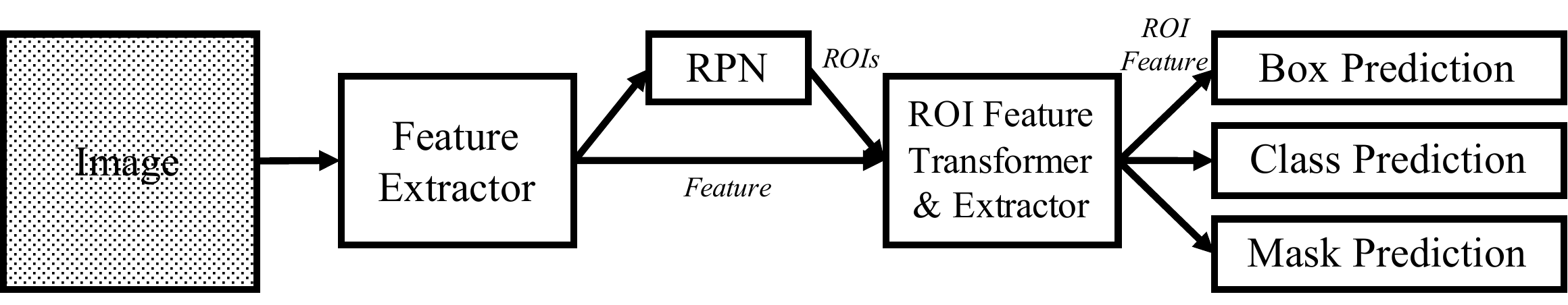}
  \caption{Instance occlusion segmentation model.}
  \label{figure:concept_of_mask_rcnn}
\end{figure}

Pixel-wise prediction in the second stage is conducted
by score regression with convolutional layers,
and softmax cross entropy is computed as the loss ($l_{mask}$) for training.
Other components are the same as the original Mask R-CNN:
\begin{itemize}
  \item {\it Feature Extractor}: For the feature extraction from input image,
    we use ResNet50-C4 (C4 represents output of 4th layer of Residual Block) \cite{He:etal:CVPR2016}
    pretrained on large-scale image classification task, ImageNet \cite{Deng:etal:CVPR2009}
    for weight initialization.
  \item {\it Region Proposal Networks (RPN)}: It is firstly proposed in \cite{Ren:etal:NIPS2015}
    for class-agonistic object bounding box prediction.
    It has two losses: bounding box regression ($l^{rpn}_{box}$) and classification
    for foreground and background ($l^{rpn}_{cls}$).
    The predicted bounding box is usually called ROI (region of interest) and
    used for the start point to predict refined class-specific box in the following components.
  \item {\it ROI Feature Transformer}: The extracted feature is transformed
    using the ROIs proposed by RPN. This normalizes the shape of the ROI
    which is important to apply instance-level reasoning to predict classes
    and refined bounding box regression by the following fully connected (FC) layers.
    ROIAlign \cite{He:etal:ICCV2017} is an operation for ROI-based feature transformation,
    and it resizes the feature with bilinear interpolation.
  \item {\it ROI Feature Extractor}: 5th layer of ResNet50 (res5) is applied to
    extract ROI-based features after the ROI feature transform.
    The weight of res5 is also copied from pretrained model in ImageNet at initialization.
  \item {\it Classification and Class-specific Box Prediction}:
    These modules are firstly proposed in \cite{Girshick:etal:ICCV2015} for multi-class object bounding box detection,
    and both are predicted by FC layers from the transformed feature for ROIs.
    Similarly to region proposal networks, it has two losses: bounding box regression ($l_{box}$)
    and multi-class classification ($l_{cls}$).
\end{itemize}
To summarize, all components are connected as shown in \figref{concept_of_mask_rcnn},
and all losses are $l^{rpn}_{box}$, $l^{rpn}_{cls}$, $l_{box}$, $l_{cls}$ and $l_{mask}$.

\begin{figure}[t]
  \setlength{\belowcaptionskip}{-10pt}
  \centering
  \begin{tabular}{c}
    \subfloat[Visible labels.]{
      \includegraphics[width=0.80\linewidth]{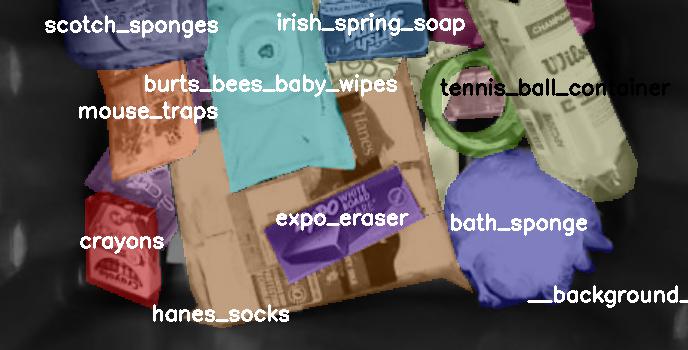}
      \figlab{semantic_inout_visible}
    }
    \\
    \subfloat[Occluded masks.]{
      \includegraphics[width=0.38\linewidth]{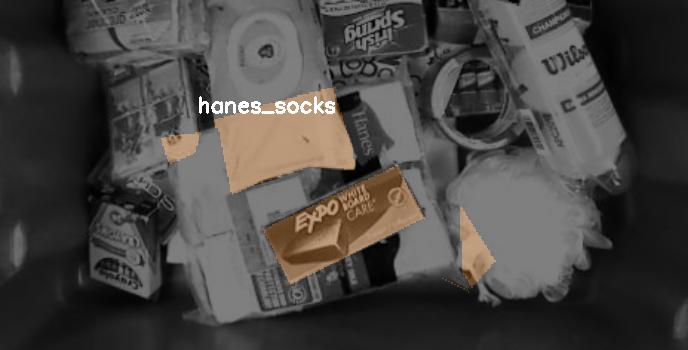}
      \includegraphics[width=0.38\linewidth]{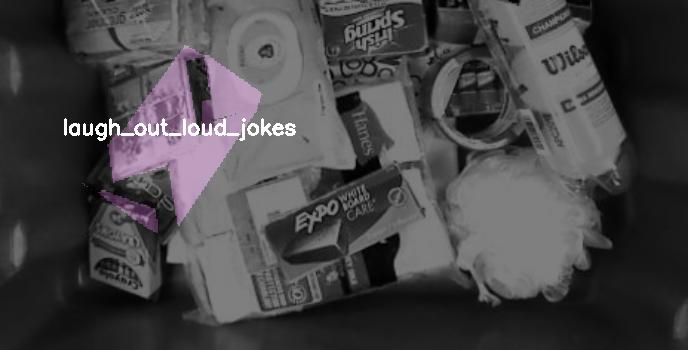}
      \figlab{semantic_inout_occluded}
    }
    % \subfloat[]{
    %   \includegraphics[width=0.26\linewidth]{semantic_visible_occluded/semantic_occluded/0004.jpg}
    %   \figlab{semantic_inout_occluded4}
    % }
  \end{tabular}
  \caption{Semantic visible and occluded labels.}
  \label{figure:semantic_inout}
\end{figure}

\subsection{Semantic Occlusion Segmentation}

For semantic occlusion segmentation, we extend previous work of semantic (visible) segmentation
by fully convolutional networks (FCN) proposed in \cite{Long:etal:CVPR2015}.
All layers are composed of convolutional or pooling layers, which keep the geometry of image,
so FCN is known as effective and widely used for pixel-wise score regression tasks:
depth prediction \cite{Eigen:etal:NIPS2014,Eigen:etal:ICCV2015},
grasp affordance \cite{Zeng:etal:ICRA2018,Zeng:etal:IROS2018,Hasegawa:etal:IROS2018},
optical flow \cite{Dosovitskiy:etal:ICCV2015}, and instance masks \cite{Dai:etal:ECCV2016,Li:etal:CVPR2017}.

FCN for semantic segmentation is composed of feature extractor and pixel-wise classification.
Original FCN \cite{Long:etal:CVPR2015} uses VGG16 \cite{Simonyan:etal:ICLR2015} as the feature extractor,
which is pretrained on image classification \cite{Deng:etal:CVPR2009}.
After the work of VGG, ResNet \cite{He:etal:CVPR2016} has been proposed for image classification,
and it showed better performance in image classification.
Following the previous work \cite{Li:etal:CVPR2017},
which uses ResNet-C4 and res5 as the feature extractor for pixel-wise score regression,
we replace the VGG feature extractor with ResNet for better feature extraction and common feature extractor as
instance segmentation in \figref{joint_learning}.

For semantic segmentation task,
pixel-wise classification module predicts $n_{class}$
object scores for each pixel.
$n_{class}$ represents number of classes including {\it background} class
that should be assigned to the other regions than the objects interest as shown in \figref{semantic_inout_visible}.
If the input RGB image has size $(H, W, 3)$, the output scores has size $(H, W, n_{class})$.
For training, softmax cross entropy loss ($l^{sem}_{vis}$) is computed for the output scores,
and at testing, the top-scored label is assigned for each pixel to get the visible label.
In occlusion segmentation, however, there can be overlaps between the occluded regions of
each class of objects. For example, in \figref{semantic_inout_occluded},
the occluded region masks of {\it hanes\_socks} and {\it laugh\_out\_loud\_jokes}
have overlap. Top-scored label assignment in semantic visible segmentation can not
handle these cases.

To handle the overlap of occluded masks, we replace the loss of pixel-wise score regression
from softmax cross entropy (class competitive) to sigmoid cross entropy (class individual):
$l^{sem}_{occ}$.
With softmax cross entropy, the model tries to find most probable class for each pixel.
On the other hand, with sigmoid cross entropy, the model tries to find the occluded probability
for each pixel individualy in all classes, which is suitable to occlusion segmentation.
Since the occluded mask is only defined for foreground,
the output size of FCN for occlusion segmentation is $(H, W, n_{class} - 1)$
(-1 represents the removal of background class).

\begin{figure}[htbp]
  \setlength{\belowcaptionskip}{-10pt}
  \centering
  \includegraphics[width=0.7\linewidth]{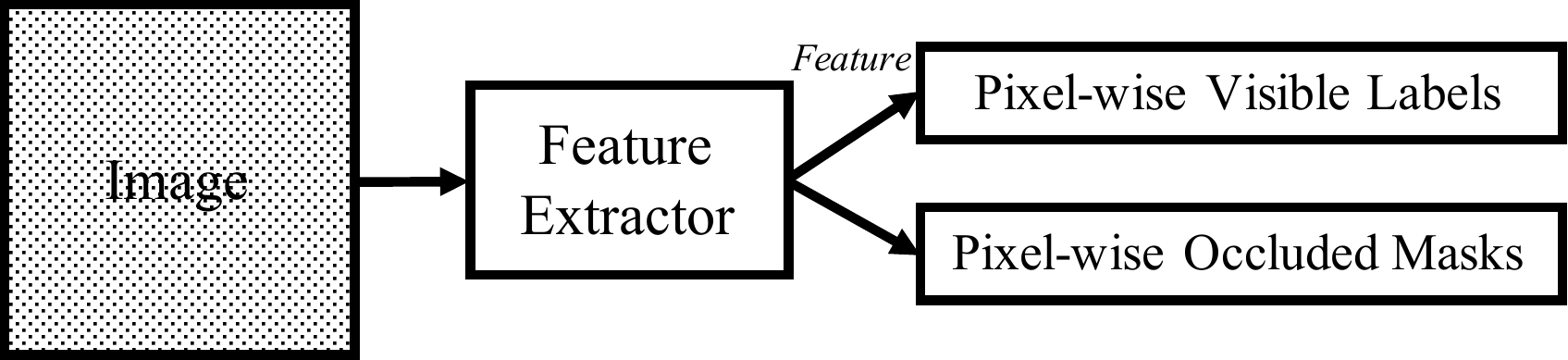}
  \caption{Semantic occlusion segmentation model.}
  \label{figure:concept_of_fcn}
\end{figure}

All components are integrated as shown in \figref{concept_of_fcn},
and all losses in semantic occlusion segmentation are
$l^{sem}_{vis}$ and $l^{sem}_{occ}$.

% -------------------------------------------------------------------------------------------------

\section{Joint Training}

\subsection{Shared Feature Extractor}

We jointly train fore-mentioned instance and semantic occlusion segmentation models.
The feature extractor which is common in the model (\figref{joint_learning})
is ResNet50-C4, and the weight of res5 is not shared.
% This is because the difference of level in feature extraction by res5
% in instance and semantic segmentation models.
In instance segmentation model, res5 ($res5^{ins}$) extractes ROI features after ROI feature transformation
from ResNet50-C4 features and ROIs. Since the ROI represents each instance in image,
this feature extraction is specific for instance (instance reasoning).
On the other hand, the res5 of semantic segmentation model ($res5^{sem}$)
extracts features in image geometry without any information about instances.

\subsection{Loss Balancing}

As described above, losses for the instance occlusion segmentation are:
\begin{itemize}
  \item $l^{rpn}_{box}$: for bounding box regression of region proposal networks (RPN);
  \item $l^{rpn}_{cls}$: for foreground vs background classification of RPN;
  \item $l^{ins}_{box}$: for instance bounding box regression;
  \item $l^{ins}_{cls}$: for instance classification;
  \item $l^{ins}_{mask}$: for instance visible, occluded and background masks.
\end{itemize}
And losses for the semantic occlusion segmentation are:
\begin{itemize}
  \item $l^{sem}_{vis}$: for pixel-wise class visible score regression;
  \item $l^{sem}_{occ}$: for pixel-wise class occluded score regression.
\end{itemize}

For joint training of both tasks, we sum all losses and backward it:
\begin{eqnarray}
  l^{ins} &=& l^{rpn}_{box} + l^{rpn}_{cls} + l^{ins}_{box} + l^{ins}_{cls} \\
  l^{sem} &=& l^{sem}_{vis} + l^{sem}_{occ} \\
  l &=& l^{ins} + \lambda \cdot l^{sem}.
\end{eqnarray}
The $\lambda$ represents the weight for loss balancing of instance and semantic occlusion segmentation.
We show experimental results by changing that parameter in the following section.

% \subsubsection{Loss Balancing}
% \subsubsection{Data Augmentation}
% Evaluation Metric

\section{Experiments}

% Evaluation Metric
% Data Augmentation

\subsection{Instance and Semantic Occlusion Segmetation}

\subsubsection{Dataset for Evaluation}

\begin{figure}[htbp]
  \vspace*{-5pt}
  \setlength{\belowcaptionskip}{-5pt}
  \centering
  \includegraphics[width=0.85\linewidth]{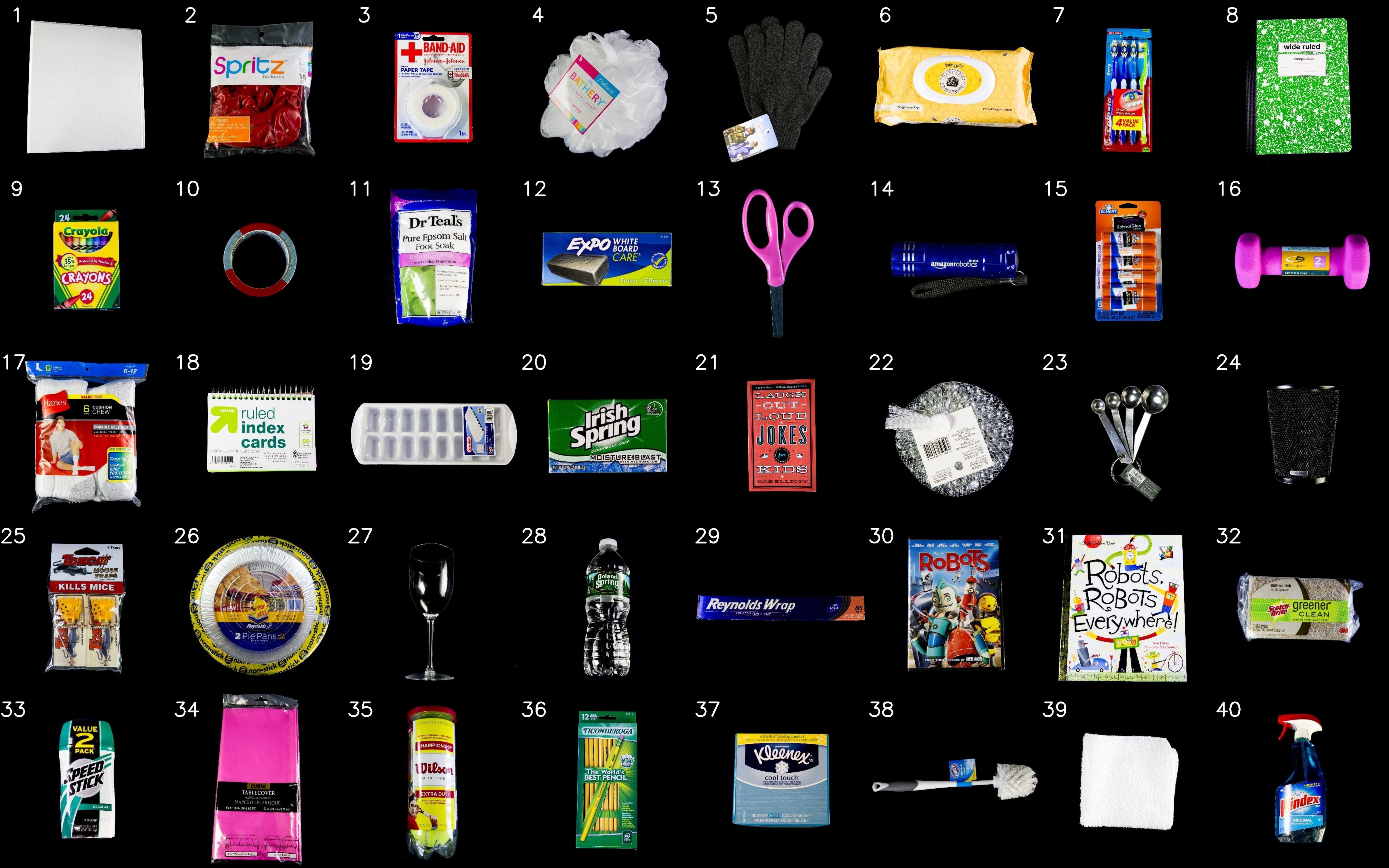}
  \caption{Objects used in the experiments.}
  \label{figure:arc2017_objects}
\end{figure}

We used 40 objects used in Amazon Robotics Challenge (ARC2017)
following our previous work \cite{Wada:etal:IROS2018} (\figref{arc2017_objects}).
For training and evaluation, we collected image frames of cluttered scene,
in addition to the dataset used in \cite{Wada:etal:IROS2018}.

\begin{figure}[htbp]
  \vspace*{-10pt}
  \setlength{\belowcaptionskip}{-5pt}
  \centering
  \begin{tabular}{c}
    \subfloat[Frame 1.]{
      \includegraphics[width=0.30\linewidth]{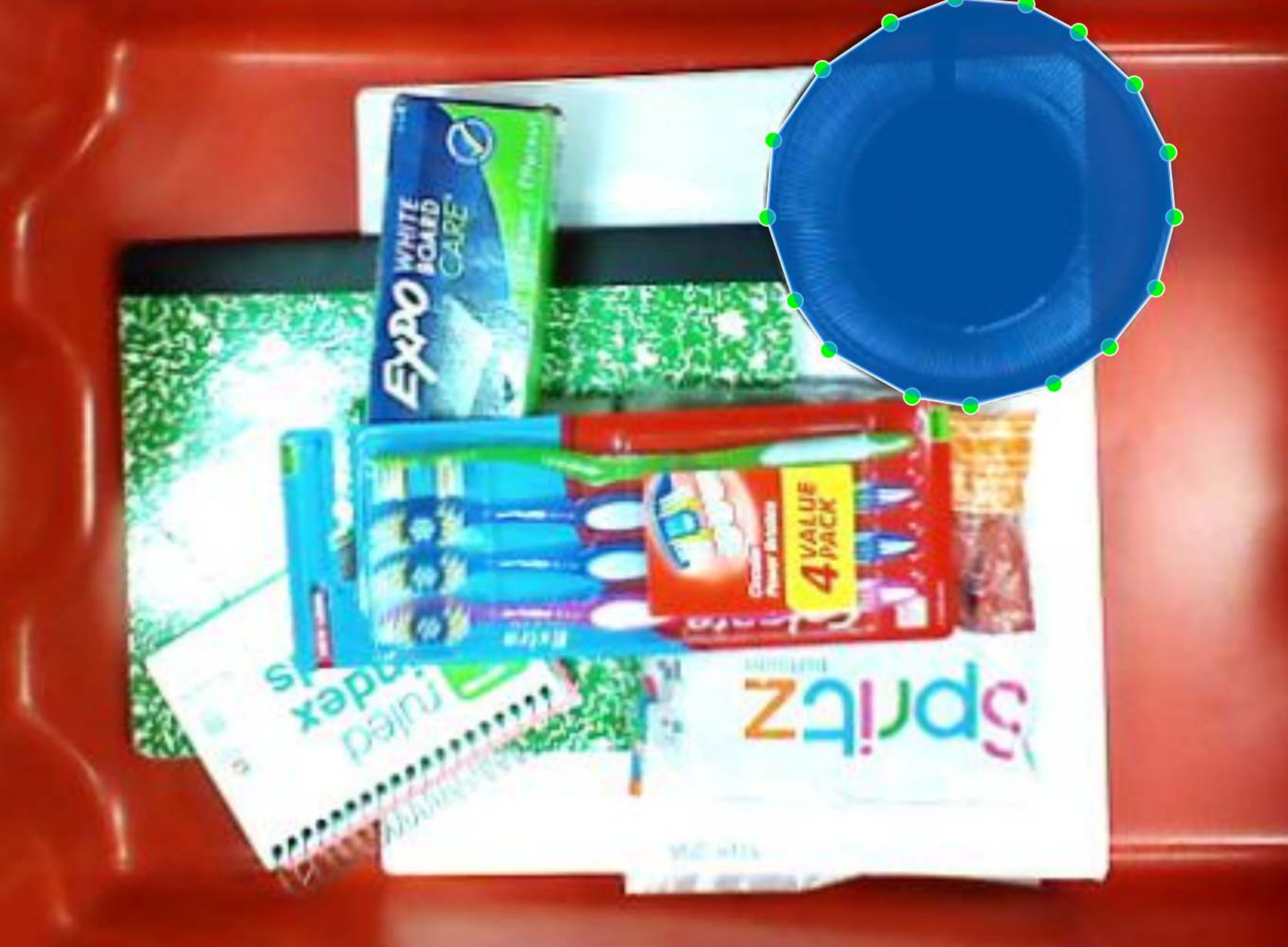}
      \figlab{dataset_creation_001}
    }
    \subfloat[Frame 2.]{
      \includegraphics[width=0.30\linewidth]{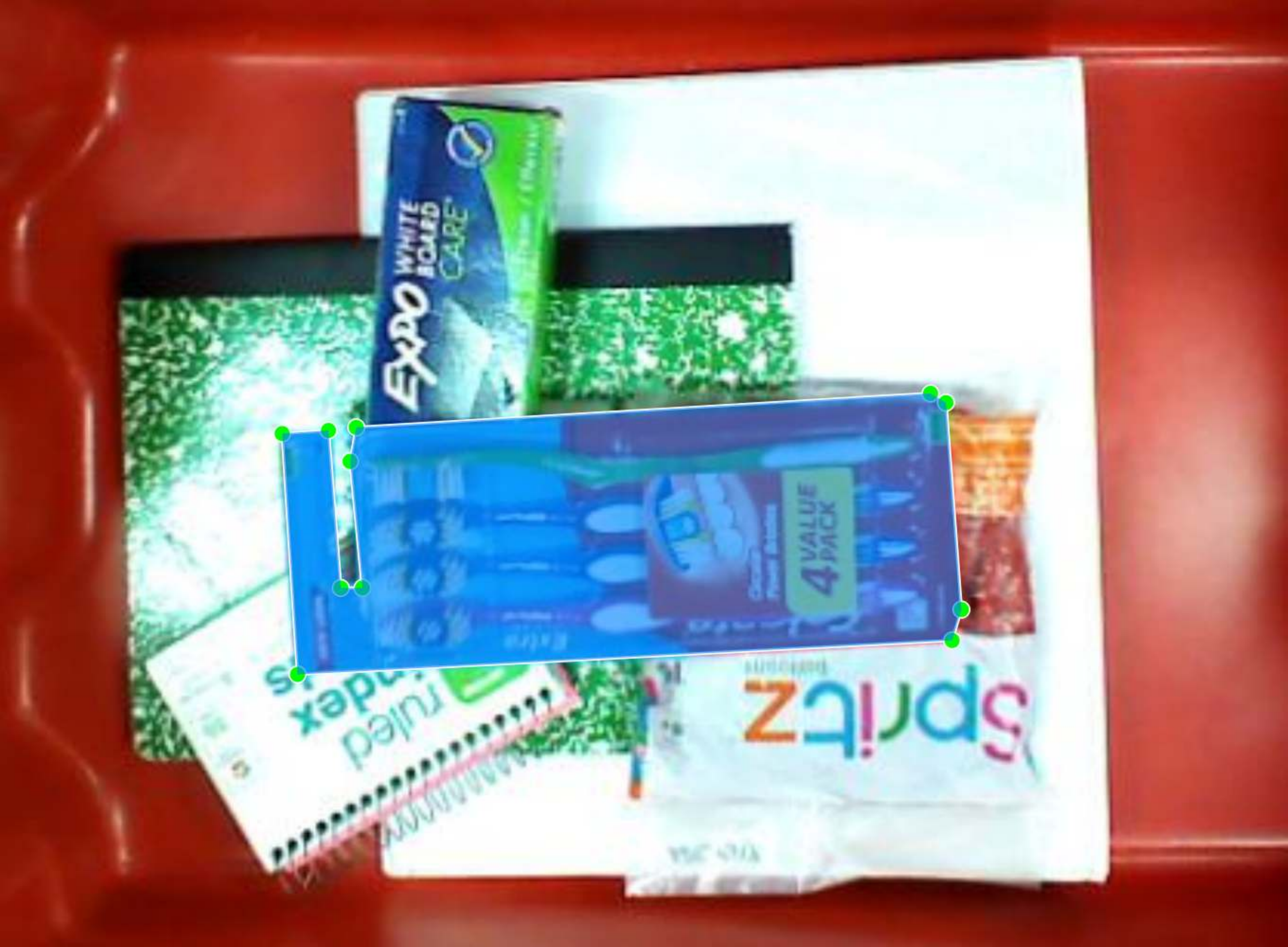}
      \figlab{dataset_creation_002}
    }
    \subfloat[Frame 3.]{
      \includegraphics[width=0.30\linewidth]{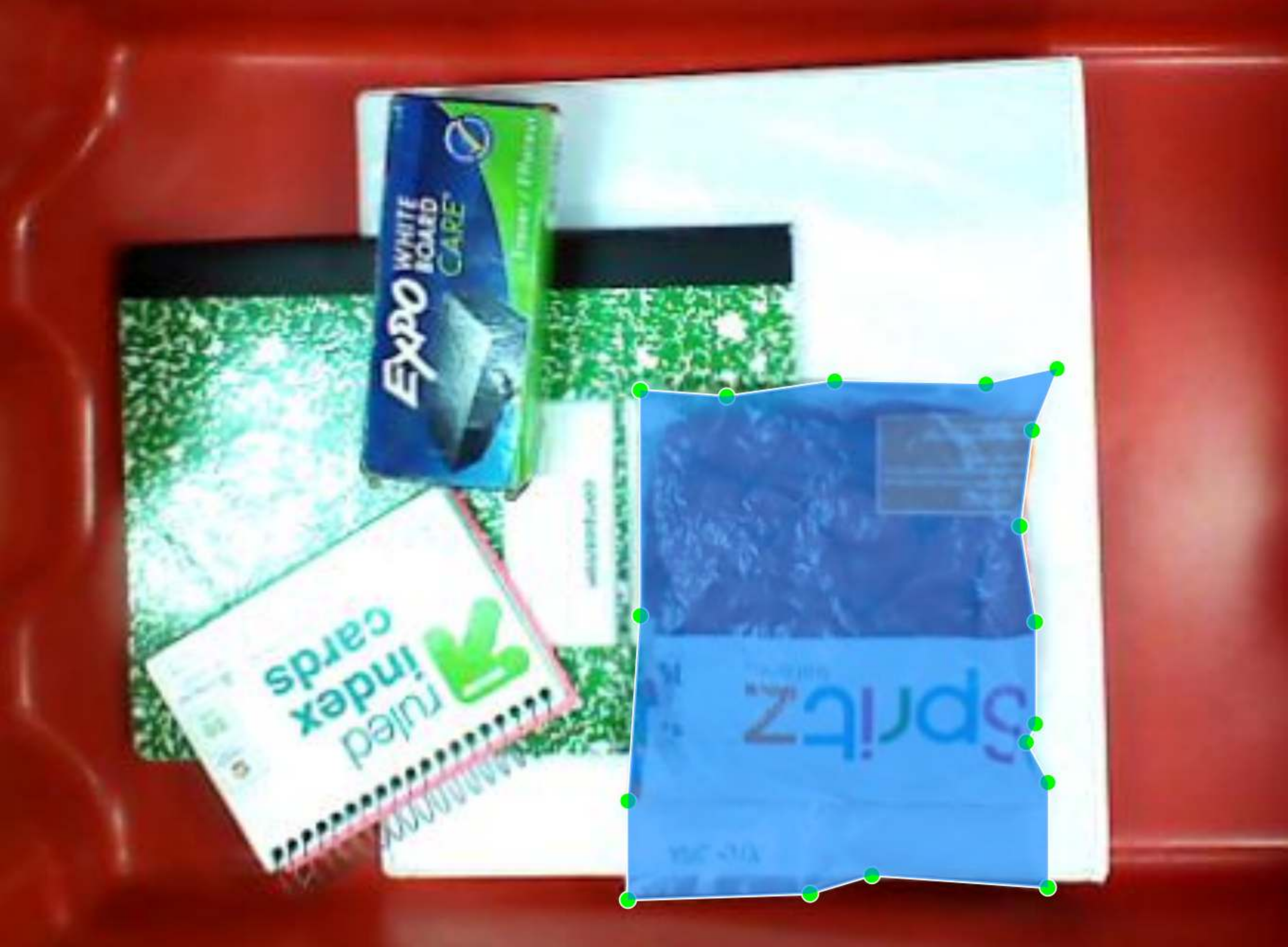}
      \figlab{dataset_creation_003}
    }
  \end{tabular}
  \caption{Annotation process in a video.}
  \label{figure:dataset_creation}
\end{figure}

For more efficient data collection and annotation, we used following annotation process:
\begin{enumerate}
  \item Fixed camera collects video frames of picking
    fully visible objects in cluttered scene by human.
  \item Human annotates the fully visible objects in the video at the frame
    where it is picked (\figref{dataset_creation}).
  \item Backproject the annotated mask to previous video frames to get
    visible and occluded masks assuming object is fixed in all frames.
\end{enumerate}
With above rules, human only needs to annotate polygon only once per each instance.
We created 51 videos (train:test = 33:18)
in addition to 22 videos (14:8) in \cite{Wada:etal:IROS2018}.
In total, there were 505 images (325:180).
The created pair of input and output are as shown in \figref{joint_learning},
with converting instance-level masks to class-level masks.

% \begin{figure}[htbp]
%   \centering
%   \includegraphics[width=\linewidth]{arc2017_objects}
%   \caption{{\bf ARC2017 Objects.}}
%   \label{figure:arc2017_objects}
% \end{figure}

\subsubsection{Evaluation Metric}

We jointly trained instance and semantic occlusion segmentation,
but the objective is improving the result of instance occlusion segmentation,
and achieving robotic picking task in scenes with heavily occluded objects.
For the evaluation, we used the metric of instance occlusion segmentation to compare the baseline models proposed in our previous work \cite{Wada:etal:IROS2018},
which is an extension of joint evaluation of detection and segmentation:
PQ = Panoptic Quality \cite{Kirillov:etal:ARXIV2018}.
Note that PQ is computed for each object class and then averaged to get the metric for multi-class
instance segmentation. It is represented as mPQ (mean of PQ).

\subsubsection{Data Augmentation}

\begin{figure}[htbp]
  \vspace*{-12pt}
  \setlength{\belowcaptionskip}{-5pt}
  \centering
  \begin{tabular}{c}
    \subfloat[Original data.]{
      \includegraphics[width=0.3\linewidth]{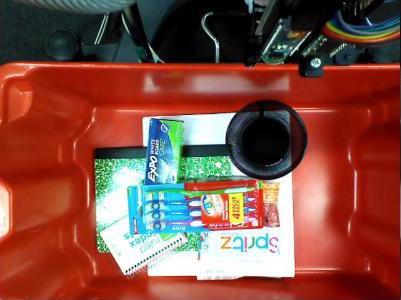}
      \includegraphics[width=0.3\linewidth]{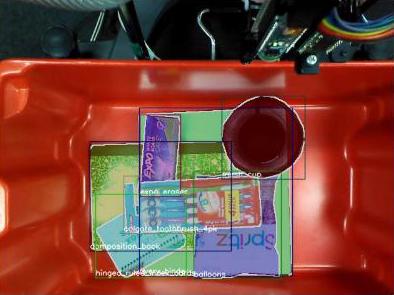}
      \includegraphics[width=0.3\linewidth]{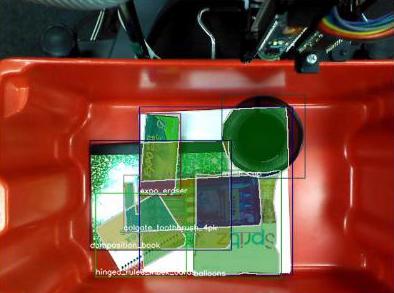}
      \figlab{data_augmentation_001}
    } \\
    \subfloat[Augmented Data.]{
      \includegraphics[width=0.3\linewidth]{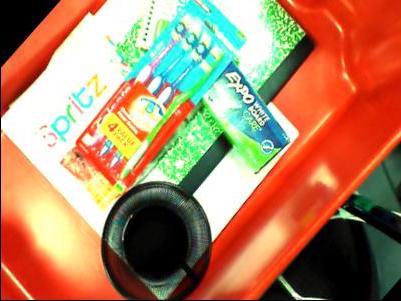}
      \includegraphics[width=0.3\linewidth]{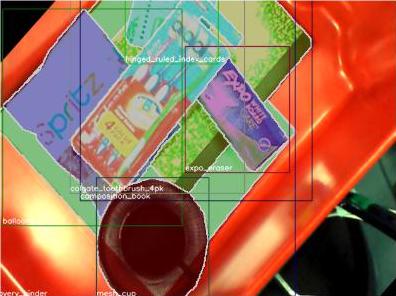}
      \includegraphics[width=0.3\linewidth]{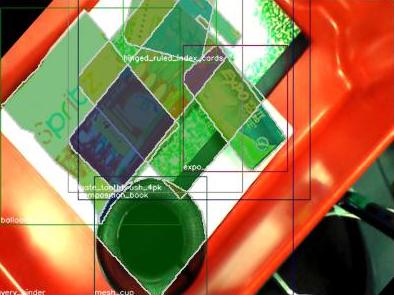}
      \figlab{data_augmentation_002}
    }
  \end{tabular}
  \caption{Data augmentation.
    {\footnotesize RGB image (right), instance visible (center) and occluded (right) masks.}
  }
  \label{figure:data_augmentation}
\end{figure}

Since objects' cluttered scene has a large number of variations even with a fixed number of objects,
data augmentation is important for robust prediction with the new images in the test dataset.
We applied following augmentations:
\begin{itemize}
  \item HSV color: for the change of brightness and the color of objects;
  \item Gaussian blur: for blur in frame by camera movement;
  \item Affine transform: for the scale, rotation, translation, and shear change.
\end{itemize}
The HSV color and Gaussian blur augmentation are applied to the RGB image of camera,
and affine transform is applied for both RGB and instance mask annotations.
The sample result of above augmentations is shown in \figref{data_augmentation}.

\subsubsection{Training Details}

We mostly followed the training parameters used in our previous work \cite{Wada:etal:IROS2018},
which slightly changes original Mask R-CNN \cite{He:etal:ICCV2017} (replacing last sigmoid to softmax).
RPN hidden channels are 512 (1024 in \cite{He:etal:ICCV2017}),
and minimum and maximum size of input image is 600 and 1000.
We used the same learning rate 0.00125 (same as \cite{He:etal:ICCV2017,Wada:etal:IROS2018}) per a batch
for training both baseline (instance-only model) and joint model.
The learning rate was multiplied by the batch size as following \cite{Goyal:etal:ARXIV2017,He:etal:ICCV2017,Wada:etal:IROS2018}.
In the following training experiments, we use the same configuration about number of GPUs (=4) and batch per gpu (=1),
so total batch size is $4 = 4 \cdot 1$ and learning rate is $0.005 = 4 \cdot 0.00125$.

\begin{table}[t]
  \setlength{\belowcaptionskip}{0pt}
  \centering
  \caption{Results of joint/non-joint learning.
    % \\
    % {\footnotesize
    %   % The model {\it instance-only} represents Mask R-CNN \cite{He:etal:ICCV2017}
    %   % with softmax for the last layer.
    %   % This result shows the joint learning helps to improve performance
    %   % of instance occlusion segmentation (mPQ).
    % }
  }
  \label{table:result_joint_learning}
  \begin{tabular}{c|c||c}
    model & $\lambda$ & mPQ\\
    \hline
    \hline
    instance-only & - & 41.0 \\
    \hline
    \multirow{5}{*}{joint (instance + semantic)} & 1 & 40.9 \\
    & 0.5 & 41.7 \\
    & \textbf{0.25} & \textbf{42.2} \\
    & 0.1 & 41.8
    % & \textbf{0.01} & \textbf{42.3} \\
  \end{tabular}
  \vspace{-4mm}
\end{table}

\subsubsection{Result}

With above training configurations,
we trained instance-only (baseline) and joint model.
For joint model, we changed the scaling hyper parameter for loss balancing
between instance $l^{ins}$ and semantic segmentation $l^{sem}$.

\tabref{result_joint_learning} shows
the result of training both instance-only (Mask R-CNN with softmax)
and joint model with using ResNet50 as the backbone of feature extractor.
It shows that the joint learning model surpasses the baseline model
the efficiency of joint learning of instance and semantic occlusion segmentation.
The different results by changing the loss balancing parameter $\lambda$ show that
the appropriate value of hyper parameter is $\lambda = 0.25$.

In order to validate the effectiveness of data augmentation,
we trained the joint model with and without the augmentation.
\tabref{results_data_augmentation} show the results and we can see
the data augmentation is fairly effective in this case.

\begin{table}[t]
  \setlength{\belowcaptionskip}{0pt}
  \centering
  \caption{Results with/without data augmentation.}
  \label{table:results_data_augmentation}
  \begin{tabular}{cccc||c}
    backbone & model & $\lambda$ & data augmentation & mPQ \\
    \hline
    \hline
    \multirow{2}{*}{ResNet50} & \multirow{2}{*}{joint} & \multirow{2}{*}{0.25} & no & 32.3 \\
    & & & \textbf{yes} & \textbf{42.2} \\
  \end{tabular}
\end{table}

We also trained with different backbone for feature extractor with replacing ResNet50 with ResNet101.
The result is shown in \tabref{result_backbones} and it shows that
the joint model outperforms the baseline model with both backbones.
\figref{qualitative_results} shows the visualization of recognition results
of joint model with ResNet101 backbone.
It shows that the capability of model to segment the occluded regions even
for the heavily occluded object: {\it avery\_binder} in \figref{qualitative_results_occ_1}.

\begin{table}[t]
  \setlength{\belowcaptionskip}{0pt}
  \centering
  \caption{Results with different backbone.}
  \label{table:result_backbones}
  \begin{tabular}{c|c|c||c}
    backbone & model & $\lambda$ & mPQ\\
    \hline
    \hline
    \multirow{2}{*}{ResNet50} & instance-only & - & 41.0 \\
    & joint & 0.25 & \textbf{42.2} \\
    \hline
    \multirow{2}{*}{ResNet101} & instance-only & - & 43.5 \\
    & joint & 0.25 & \textbf{44.5}
  \end{tabular}
  \vspace{-4mm}
\end{table}

\begin{figure}[htbp]
  \vspace*{-15pt}
  \setlength{\belowcaptionskip}{-8pt}
  \centering
  \begin{tabular}{c}
    \subfloat[Image.]{
      \includegraphics[width=0.15\textwidth]{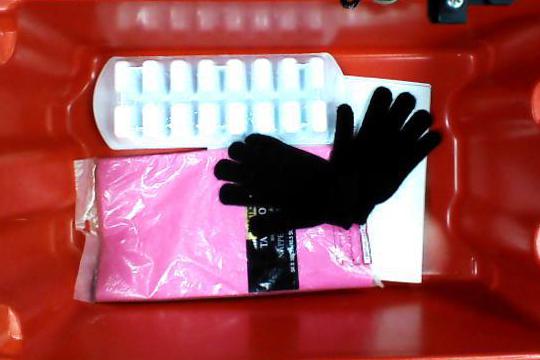}
      \figlab{qualitative_results_image}
    }
    \subfloat[Semantic segmentation.]{
      \includegraphics[width=0.15\textwidth]{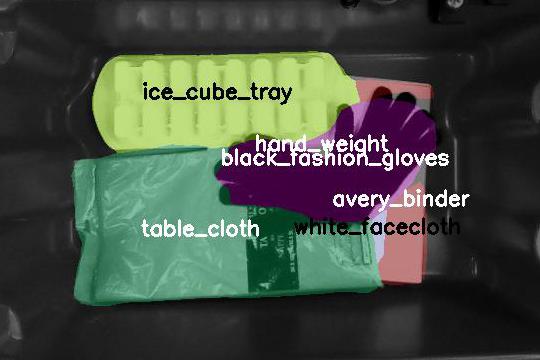}
      \figlab{qualitative_results_sem_vis}
    }
    \subfloat[Instance masks of {\it avery\_binder} (1).]{
      \includegraphics[width=0.15\textwidth]{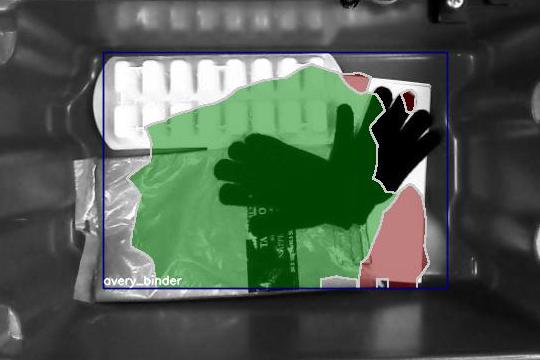}
      \figlab{qualitative_results_occ_0}
    } \\
    \subfloat[Instance masks of {\it black\_fashion\_gloves} (5).]{
      \includegraphics[width=0.15\textwidth]{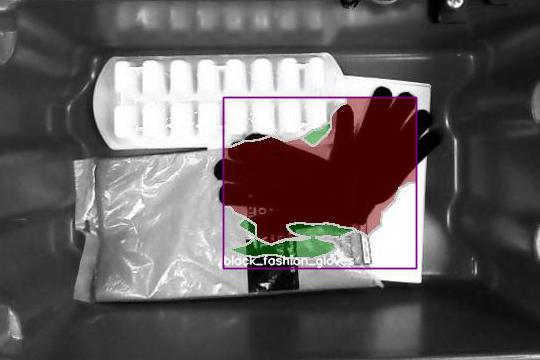}
      \figlab{qualitative_results_occ_1}
    }
    \subfloat[Instance masks of {\it ice\_cube\_tray} (19).]{
      \includegraphics[width=0.15\textwidth]{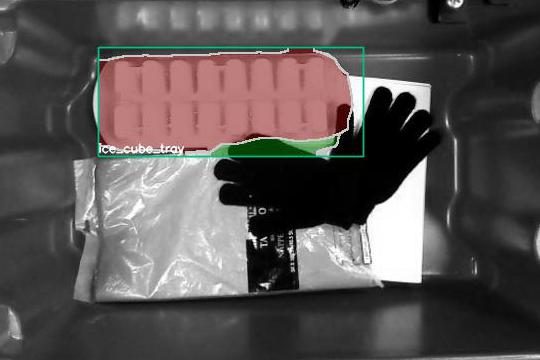}
      \figlab{qualitative_results_occ_2}
    }
    \subfloat[Instance masks of {\it table\_cloth} (34).]{
      \includegraphics[width=0.15\textwidth]{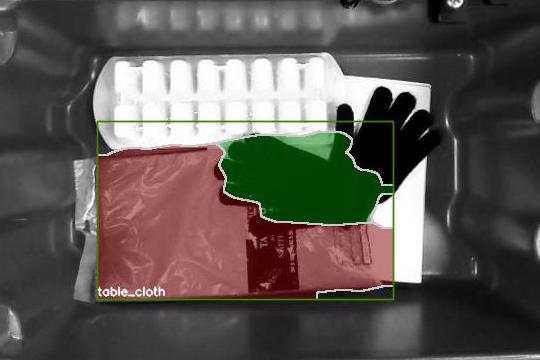}
      \figlab{qualitative_results_occ_3}
    }
  \end{tabular}
  \caption{Qualitative results.
    \footnotesize{The number after object name corresponds to the number in \figref{arc2017_objects}.
      Red mask represents \textcolor{red}{visible}, and green mask represents \textcolor{Green}{occluded}.
    }
  }
  \label{figure:qualitative_results}
\end{figure}

% -------------------------------------------------------------------------------------------------

\subsection{Robotic Pick-and-Place Experiments}

We evaluated the proposed model in the 2 types of robotic pick-and-place tasks:
\begin{itemize}
  \item Random Picking: in which robot is requested to move all objects from one to another,
    without any priority of the order.
  \item Target Picking: in which robot is requested to pick a designated object,
    with removing the obstacle object appripriately.
\end{itemize}
Even in the target picking, random picking strategy can be also used, however,
picking without any priority takes time to reach to the target object.

\begin{figure*}[t]
  \centering
  \subfloat[]{
    \includegraphics[width=0.15\linewidth]{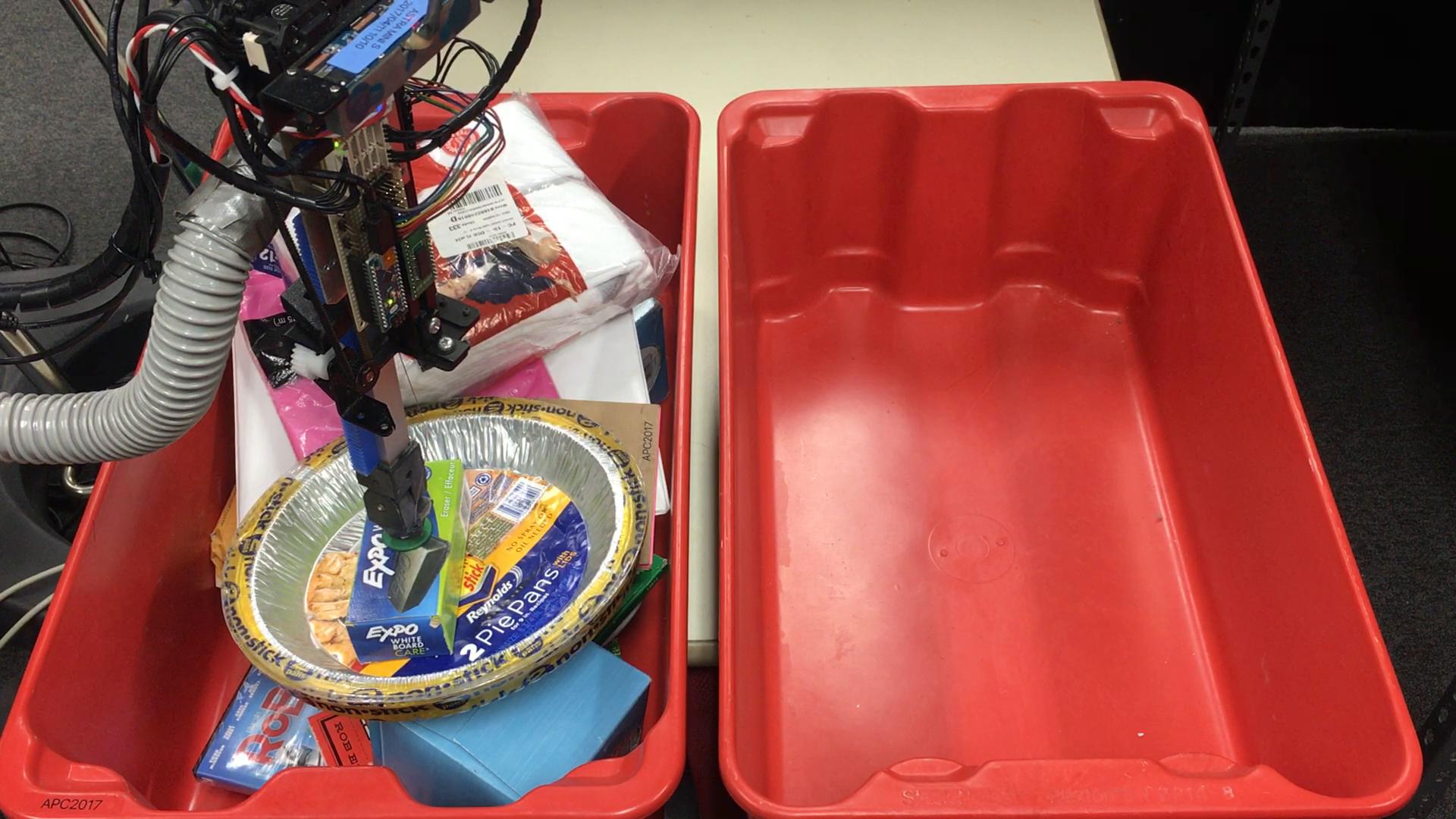}
    \figlab{random_picking_1}
  }
  \subfloat[]{
    \includegraphics[width=0.15\linewidth]{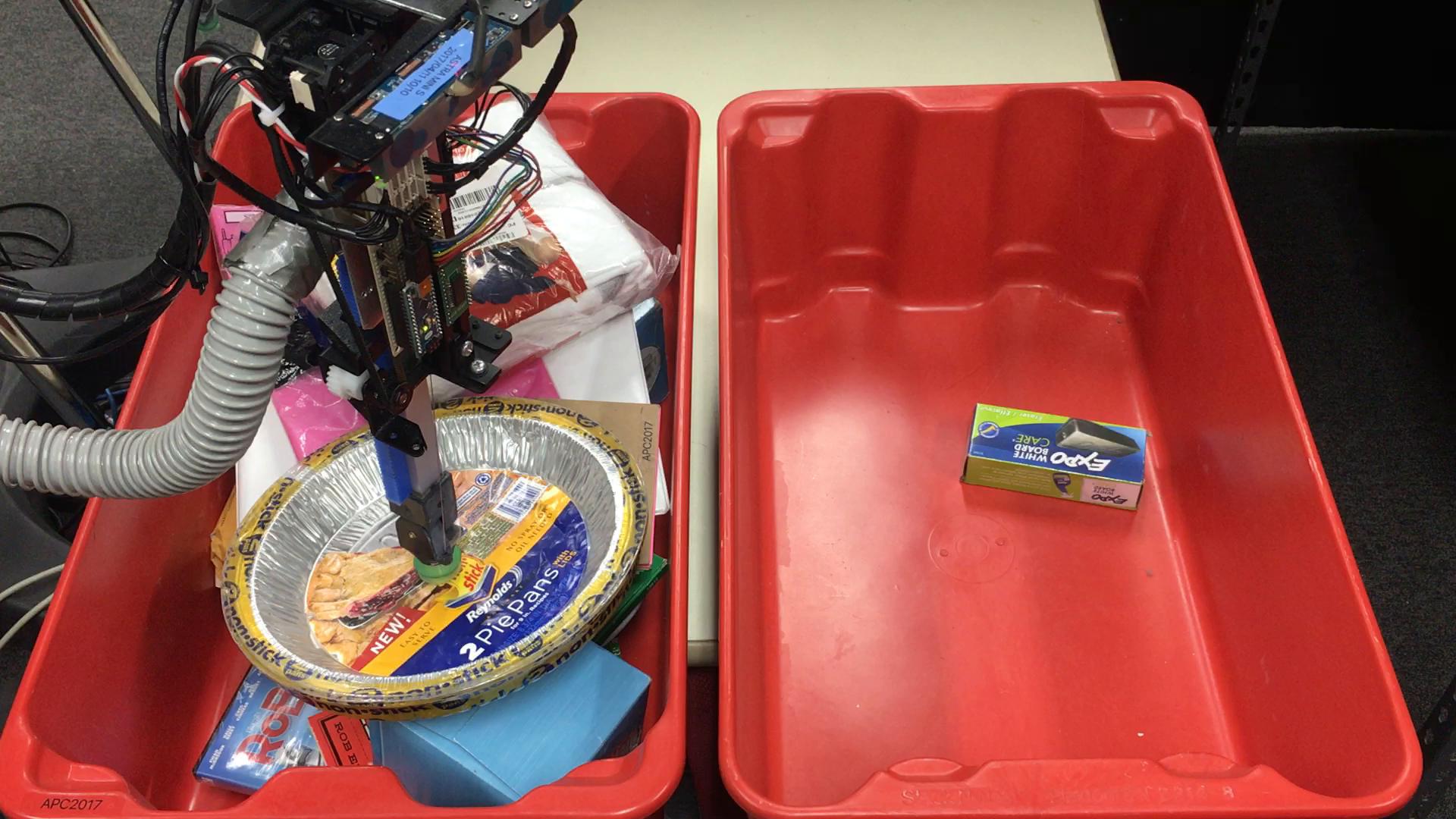}
    \figlab{random_picking_2}
  }
  \subfloat[]{
    \includegraphics[width=0.15\linewidth]{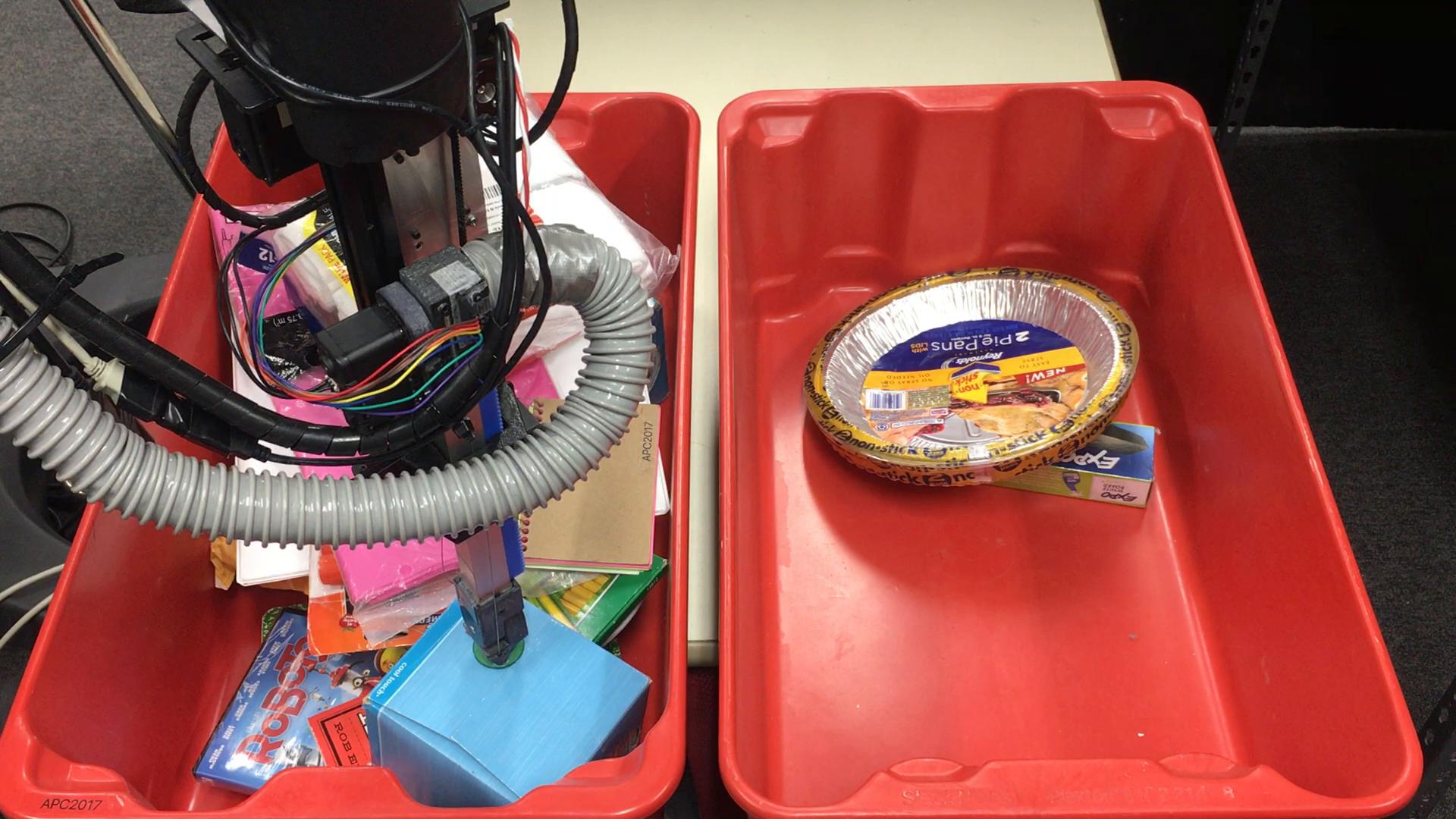}
    \figlab{random_picking_3}
  }
  \subfloat[]{
    \includegraphics[width=0.15\linewidth]{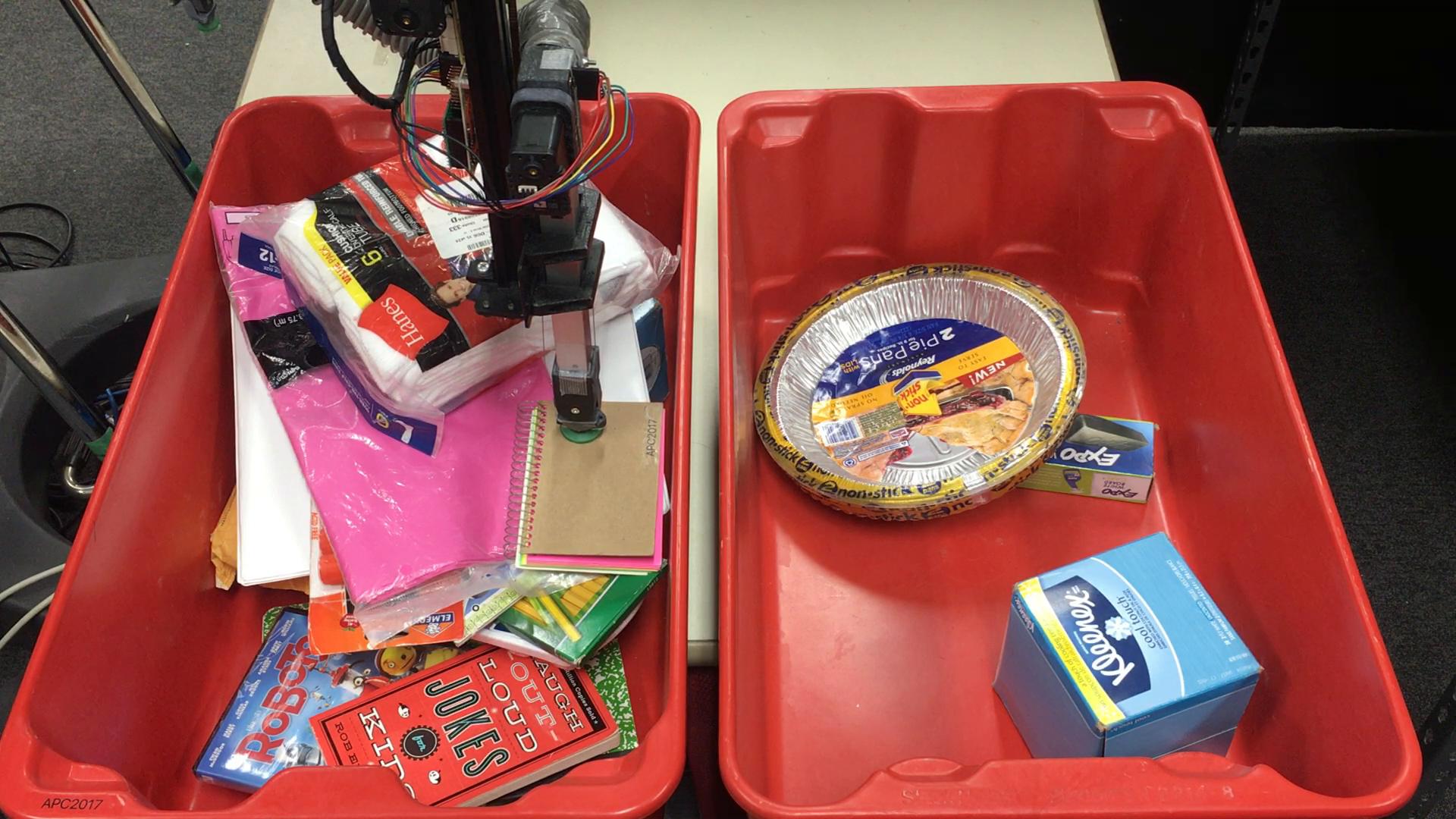}
    \figlab{random_picking_4}
  }
  \subfloat[]{
    \includegraphics[width=0.15\linewidth]{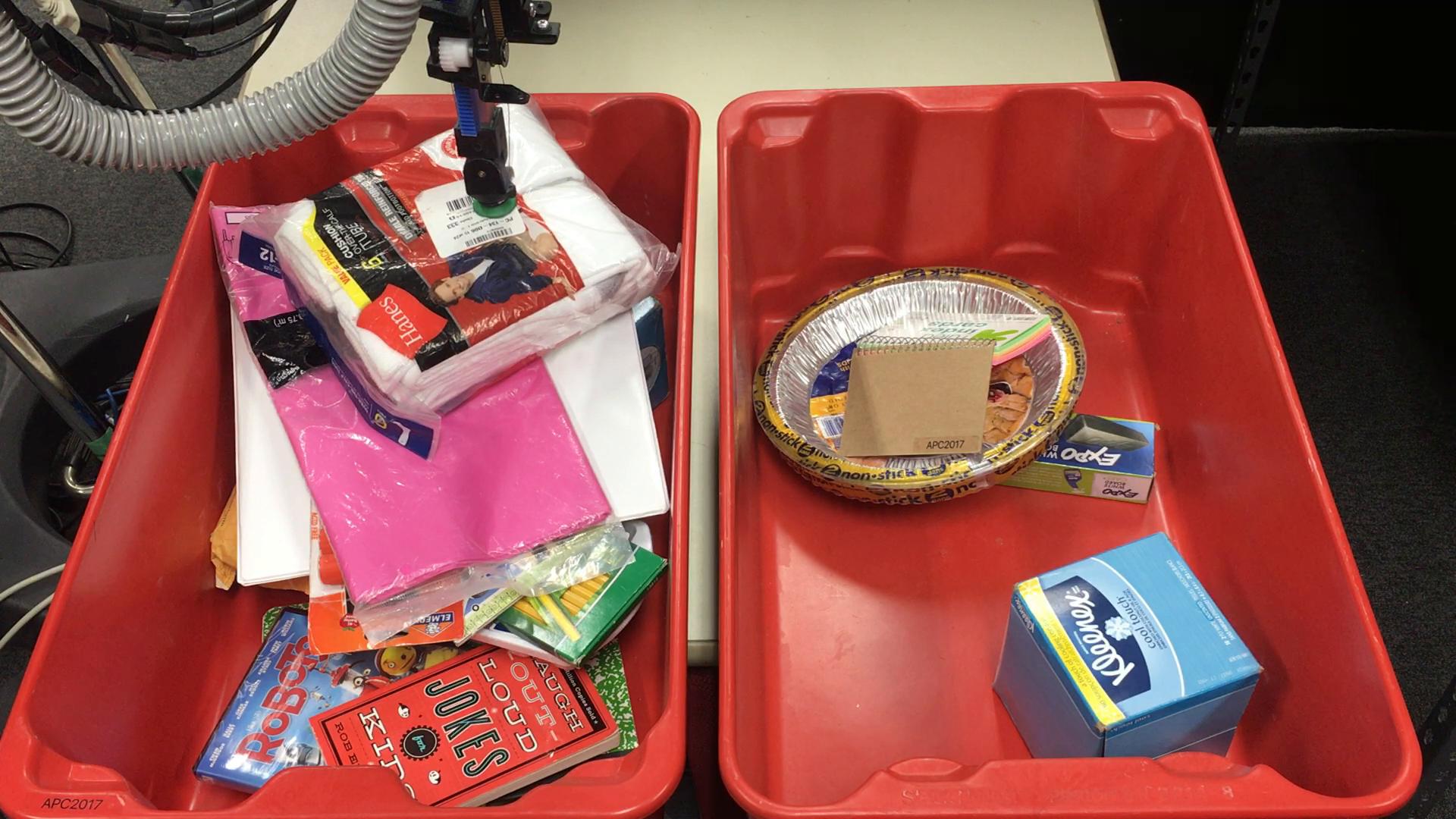}
    \figlab{random_picking_5}
  }
  \subfloat[]{
    \includegraphics[width=0.15\linewidth]{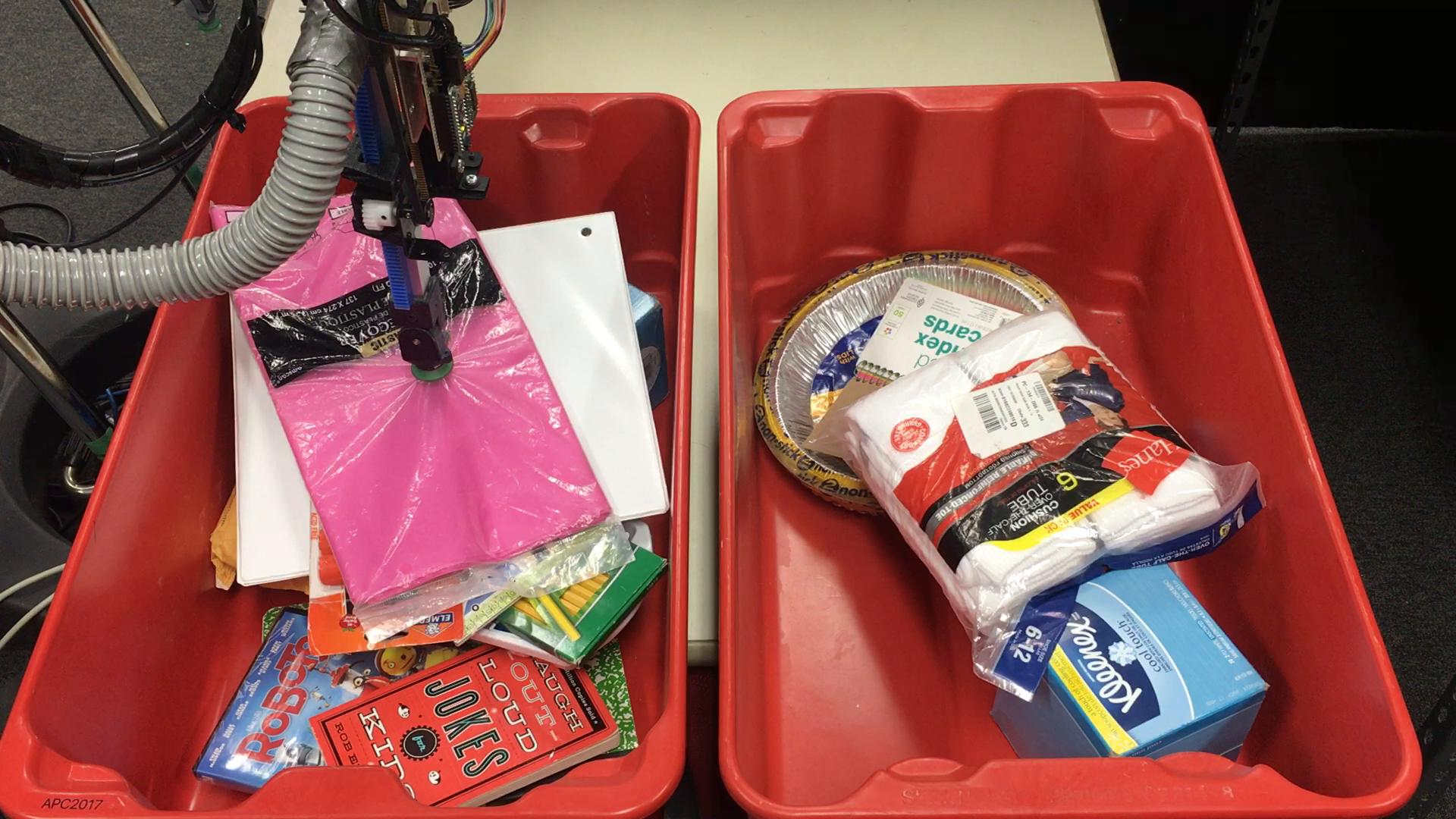}
    \figlab{random_picking_6}
  }
  \caption{Random picking.
    \footnotesize{
      The robot is requested to pick and place
      of all objects located in left bin (source) to the right bin (destination).
    }
  }
  \label{figure:random_picking}
\end{figure*}

\begin{figure*}[t]
  \centering
  \subfloat[]{
    \includegraphics[width=0.15\linewidth]{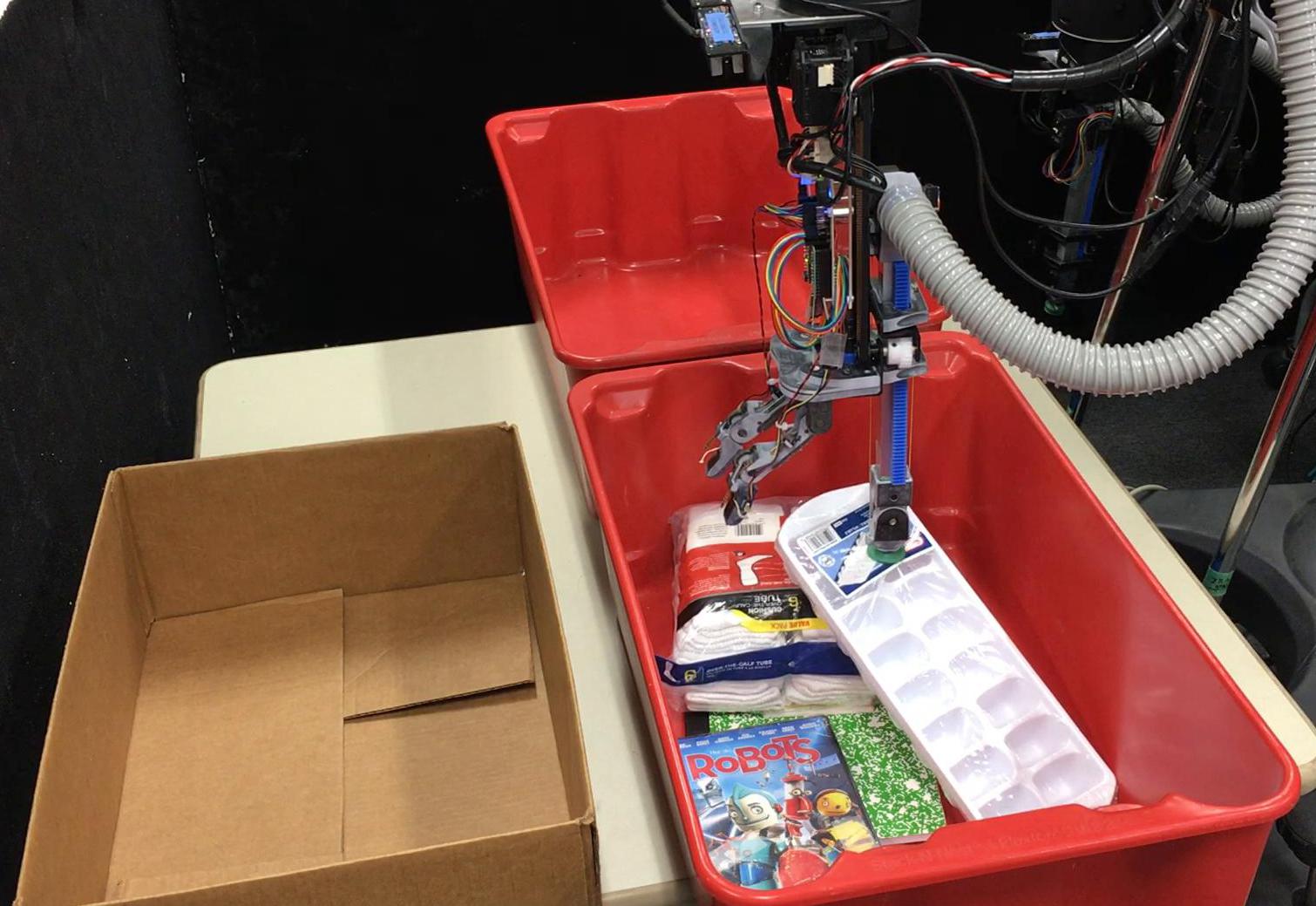}
    \figlab{target_picking_1}
  }
  \subfloat[]{
    \includegraphics[width=0.15\linewidth]{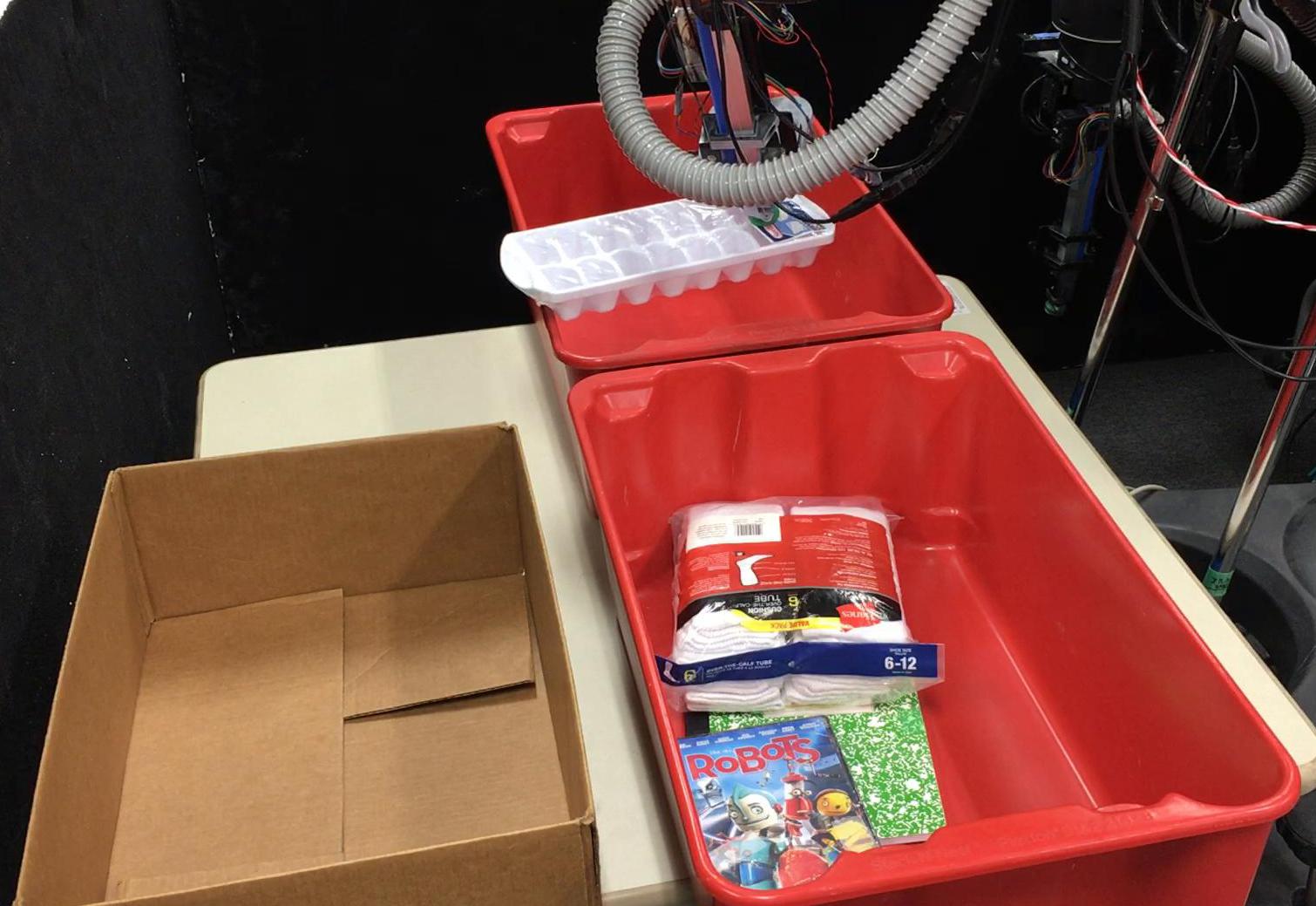}
    \figlab{target_picking_2}
  }
  \subfloat[]{
    \includegraphics[width=0.15\linewidth]{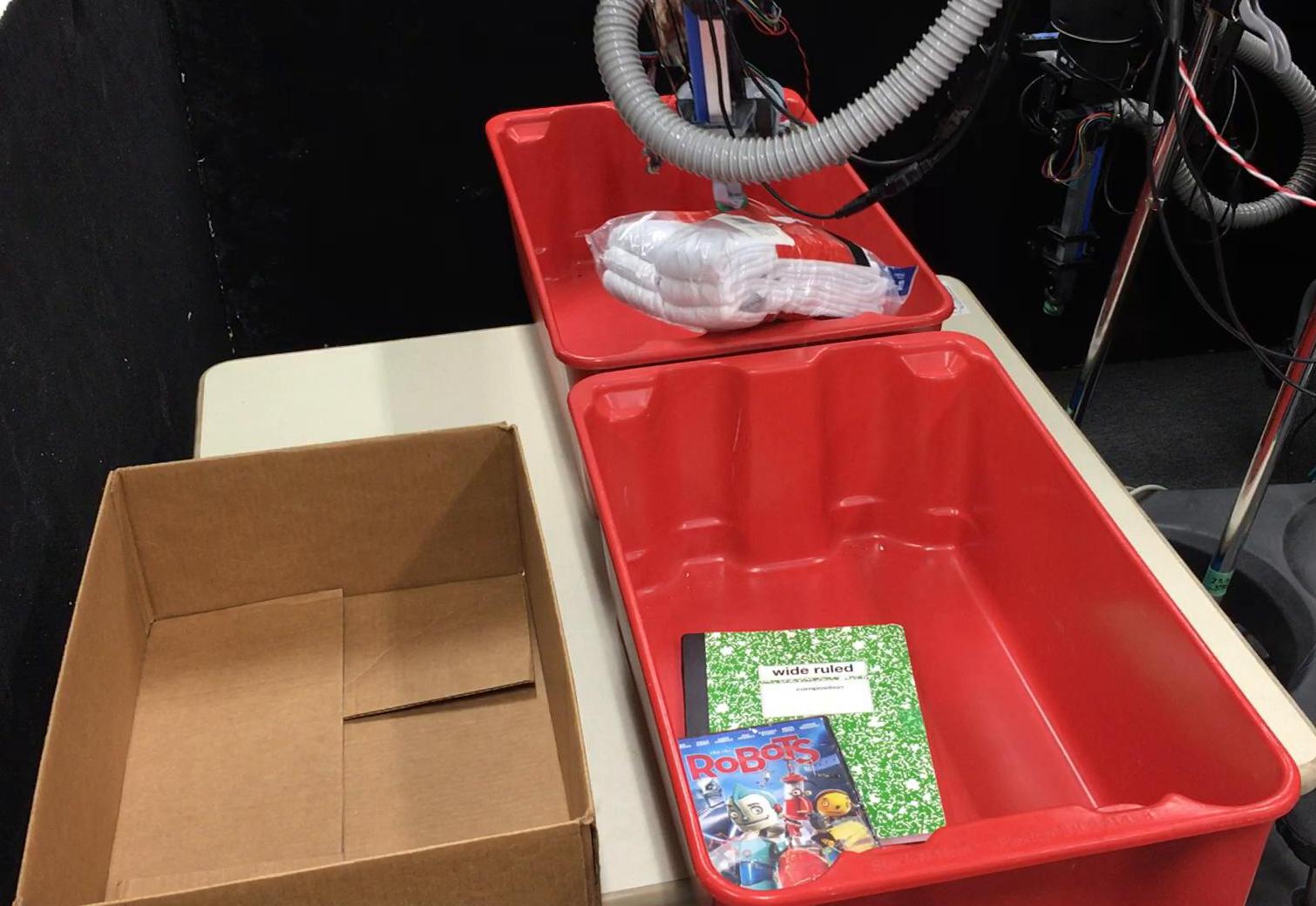}
    \figlab{target_picking_3}
  }
  \subfloat[]{
    \includegraphics[width=0.15\linewidth]{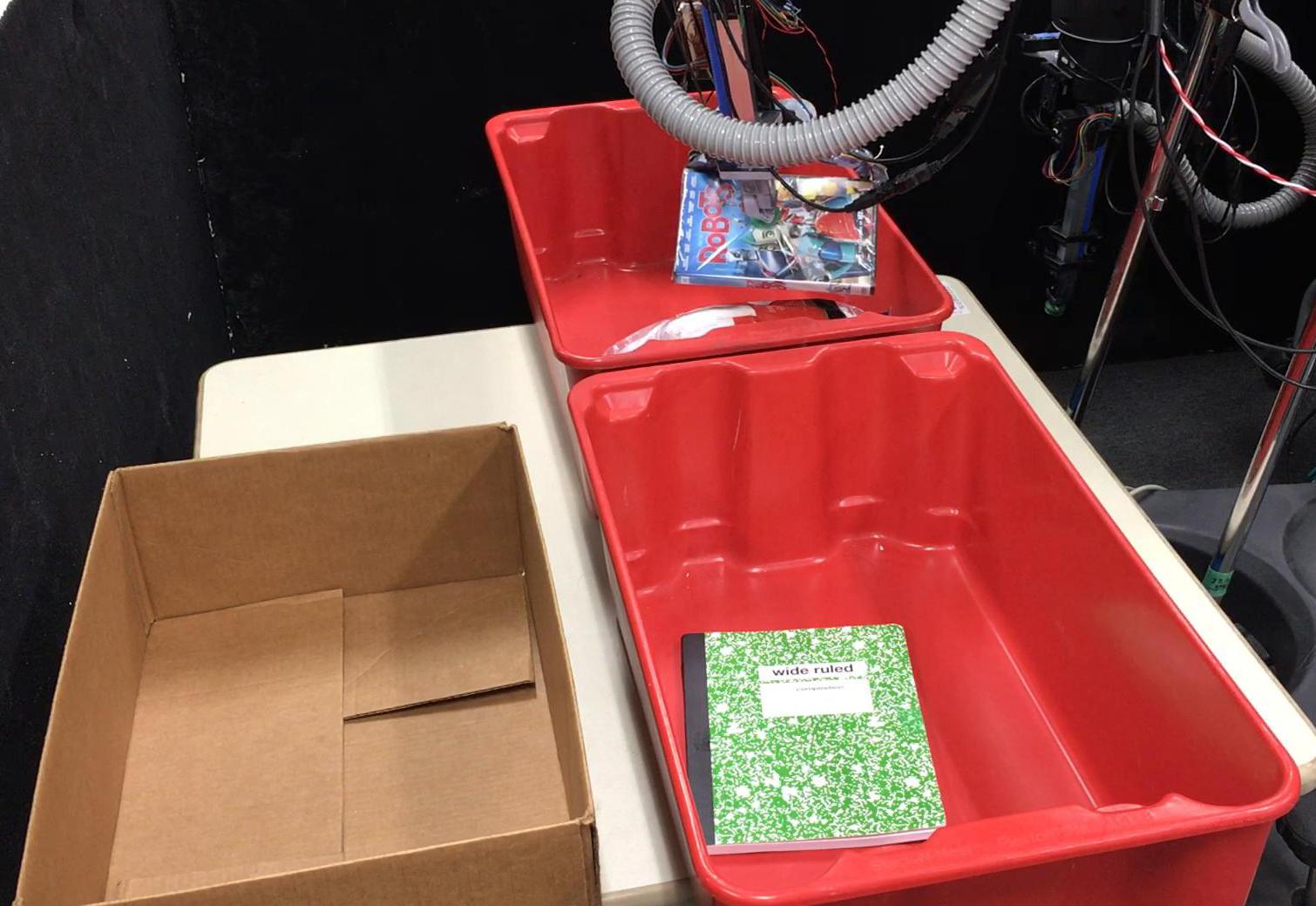}
    \figlab{target_picking_4}
  }
  \subfloat[]{
    \includegraphics[width=0.15\linewidth]{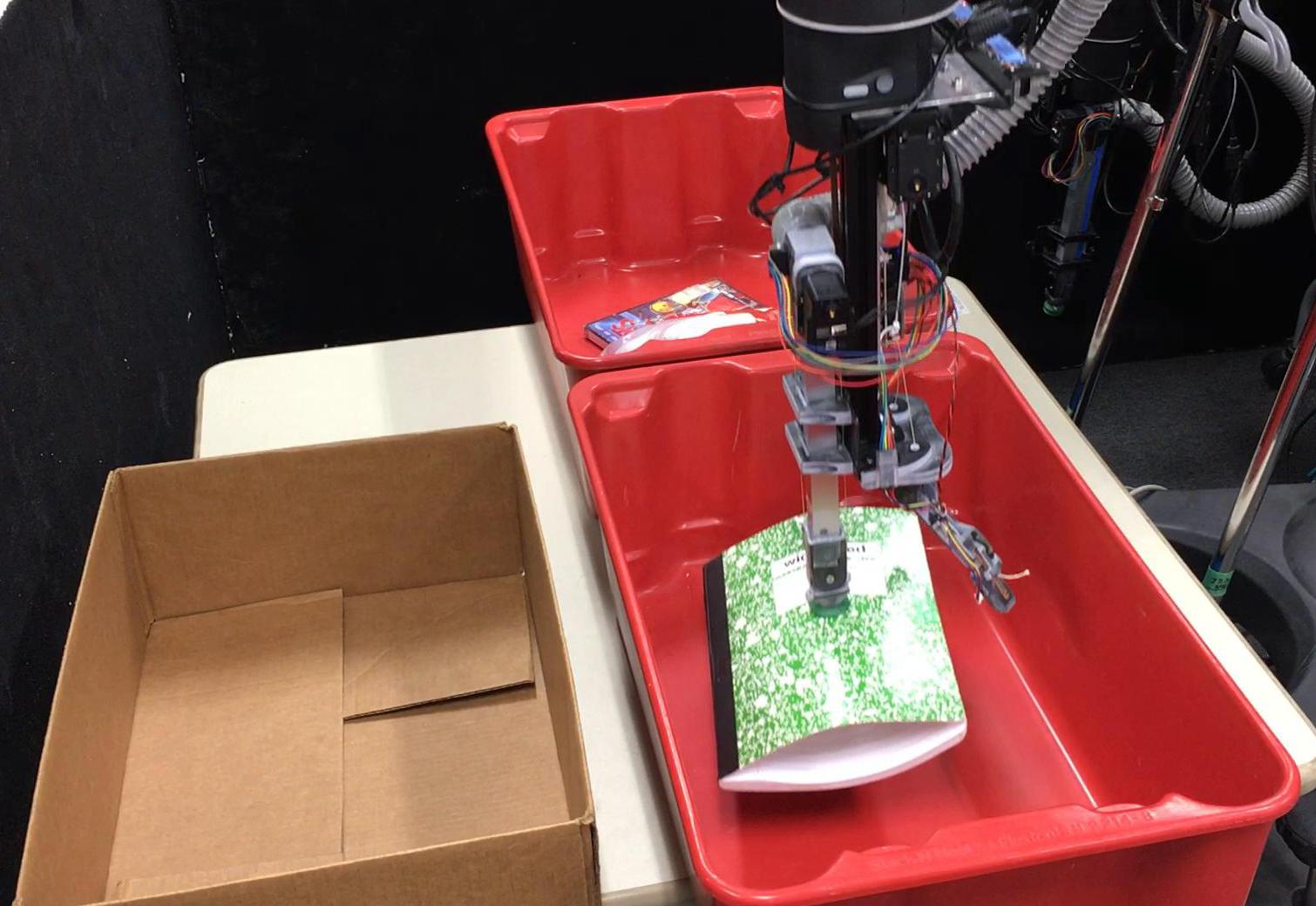}
    \figlab{target_picking_5}
  }
  \subfloat[]{
    \includegraphics[width=0.15\linewidth]{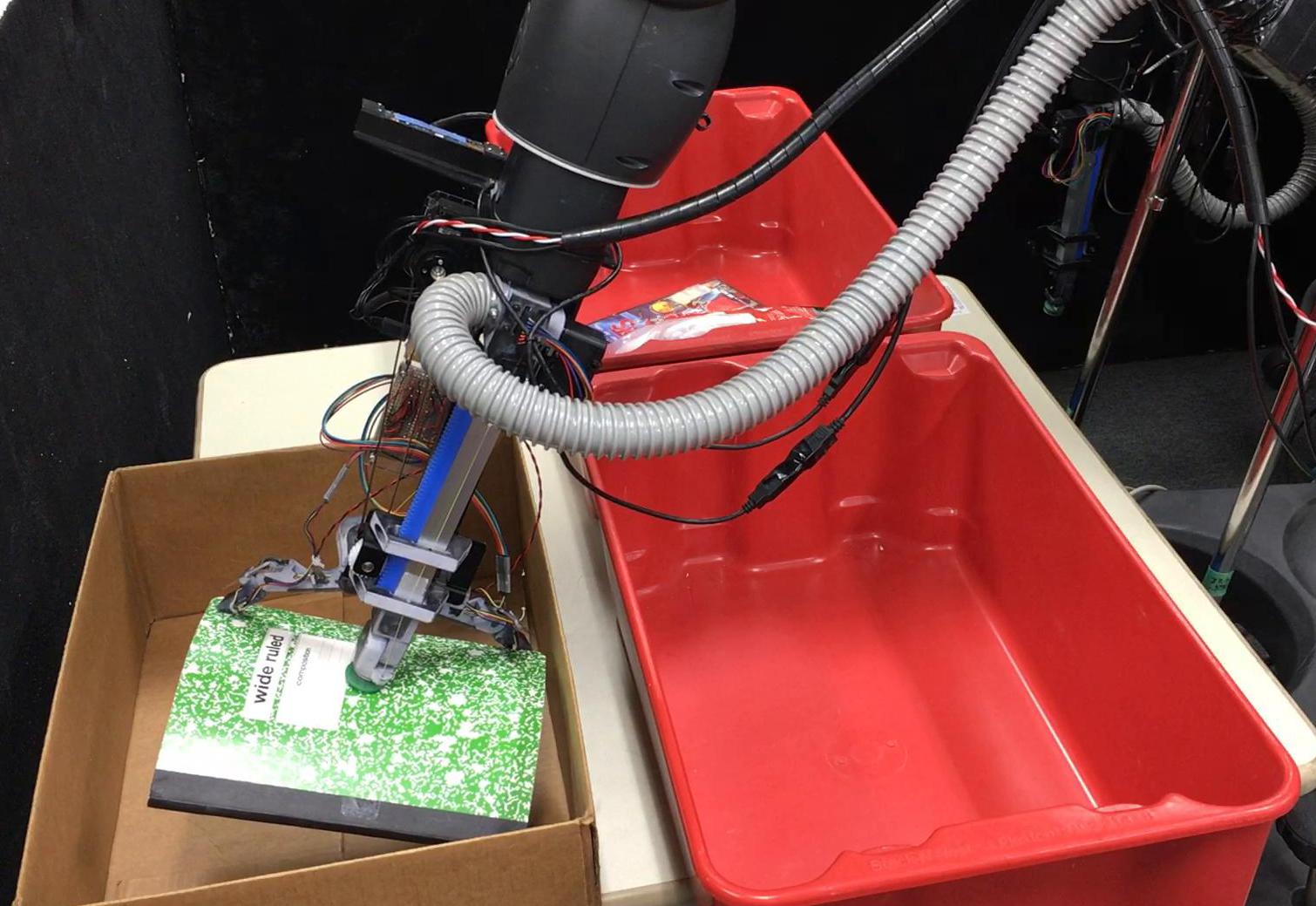}
    \figlab{target_picking_6}
  }
  \caption{Target picking.
    {\footnotesize
      The robot is requested to move a heavily occluded target object (green book)
      into the cardboard box.
      Since the target object is occluded by other objects (ice cube tray, socks),
      the robot needs to detect and remove these obstacle objects into another bin
      to reach the target object.
    }
  }
  \label{figure:target_picking}
\end{figure*}

\subsubsection{Random Picking}

In random picking, there are 2 typical failure cases:
\begin{itemize}
  \item Next target object is occluded by other heavy objects and can not be
    picked because of the collision.
  \item Next target object is occluded by other light objects and robot mistakenly picks
    2 or more objects at once.
\end{itemize}
In general, the former case causes picking failure, time loss and damage to items,
and latter case causes dropping one of the picked objects,
and wrong count of the picking.
To avoid the above problems in random picking, we used occlusion segmentation results
to select not-occluded objects as the appropriate next target.

% \begin{figure}[htbp]
%   \centering
%   \includegraphics[width=0.9\linewidth]{random_picking_ws}
%   \caption{{\bf Workspace of random picking.}}
%   \label{figure:random_picking_ws}
% \end{figure}

We used 23 objects in \figref{arc2017_objects} which can be graspable by suction
for random picking evaluation.
The task is moving objets from source (left bin) and destination (right bin)
in \figref{random_picking}.
We used the suction gripper we developed before \cite{Hasegawa:etal:IROS2017},
and used the centroid of point cloud extracted by the visible region segmentation
for the suction point.

We randomly created cluttered scenes in the bin,
and experimented with the random picking by robot based on the occluded region segmentation.
In 67 attempts (1 attempt means picking one object) of pick-and-place, the robot:
\begin{itemize}
  \item successfully picked an object in 63 times (94.0\%);
  \item failed to grasp because of wrong segmentation of visible regions in 2 times (3\%);
  \item failed to pick because of collision of other object once (1.5\%);
  \item mistakenly picked 2 objects at the same time because of wrong segmentation of
    occluded regions in once (1.5\%).
\end{itemize}
The result shows the effectiveness of the model
to select the fully visible (not-occluded) object in the random picking.

\subsubsection{Target Picking}

In target picking, typical failure case is
that the target object is so heavily occluded that the robot cannot find it.
In this case, the robot can shift to random picking, however,
if the target object can be detected even with heavy occlusion,
it is useful to plan the appropriate picking order.

\figref{target_picking_recog_rgb} shows the typical difficult case of target picking,
in which the green book (8 in \figref{arc2017_objects}) is located under other 3 objects.
\figref{target_picking_recog_vis}, \ref{figure:target_picking_recog_occ}
show the visible and occluded mask of each object,
in which the same object region is visualized with the same color.
Red region represents the masks for the green book, and it shows the model successfully segmented it.

\begin{figure}[htbp]
  \centering
  \subfloat[Image.]{
    \includegraphics[width=0.15\textwidth]{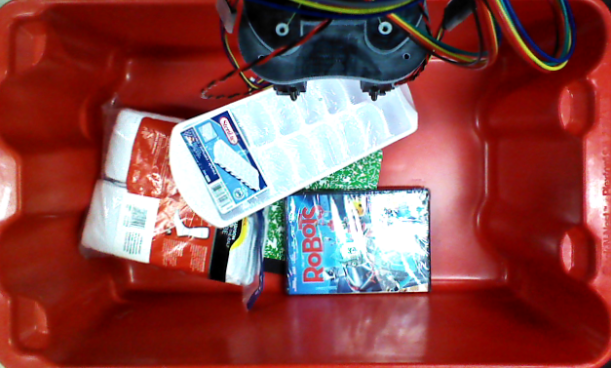}
    \figlab{target_picking_recog_rgb}
  }
  \subfloat[Visible masks.]{
    \includegraphics[width=0.15\textwidth]{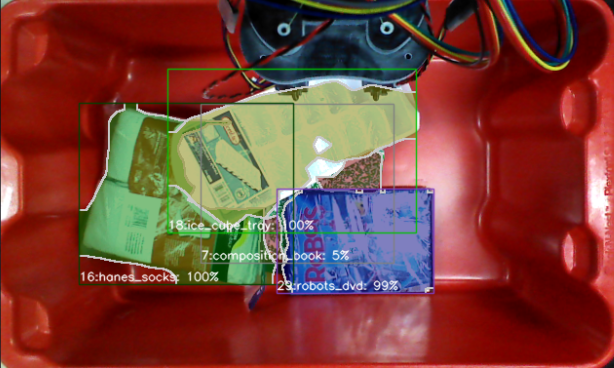}
    \figlab{target_picking_recog_vis}
  }
  \subfloat[Occluded masks.]{
    \includegraphics[width=0.15\textwidth]{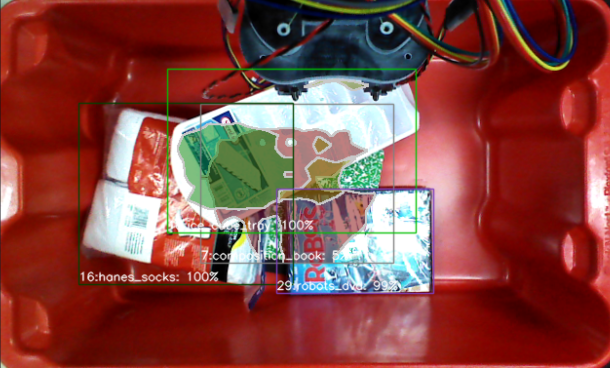}
    \figlab{target_picking_recog_occ}
  }
  \caption{Typical difficult case of target picking.}
  \label{figure:target_picking_recog}
\end{figure}

\figref{target_picking} shows the demonstration of picking
the heavily occluded green book.
The bin in closer side is the source location,
bin in the further side is the destination for obstacle objects (non target),
and the cardboard box is the destination for the target object.
This demonstration shows the effectiveness of the model
in target picking task with heavily occluded targets.

\section{Conclusions}

We presented the joint learning of instance and semantic segmentation
especially for instance-level visible and occluded region segmentation.
For collaboration of semantic segmentation with instance occlusion segmentation,
we introduced semantic occlusion segmentation
extending the conventional semantic visible segmentation.
The experimental results showed the effectiveness of the joint training
comparing with only training instance segmentation.
We also evaluated the model in 2 robotic pick-and-place tasks (random and target picking),
and showed the effectiveness in picking tasks of various objects.

%                                   % the textheight of the last page by a suitable amount.
%                                   % This command does not take effect until the next page
%                                   % so it should come on the page before the last. Make
%                                   % sure that you do not shorten the textheight too much.

\bibliographystyle{junsrt}
\bibliography{main}

\end{document}